\documentclass[journal,twoside,web]{ieeecolor}
\usepackage{tmi}
\usepackage{cite}
\usepackage{amsmath,amssymb,amsfonts}
\usepackage[ruled,vlined,linesnumbered]{algorithm2e}
\usepackage[caption=false,font=footnotesize]{subfig}
\usepackage[hidelinks]{hyperref}
\usepackage{cleveref}
\usepackage{graphicx}
\usepackage{float}
\usepackage{multirow}
\usepackage{booktabs}
\let\labelindent\relax
\usepackage{enumitem}

\usepackage{xcolor}
\usepackage{adjustbox}

\newcommand{\rrevise}[1]{\textcolor{black}{#1}}
\newcommand{\myrevise}[1]{\textcolor{black}{#1}}
\newcommand{\revision}[1]{\textcolor{black}{#1}}
\usepackage{booktabs}
\usepackage{threeparttable}
\newcommand{\sym}[1]{\textsuperscript{#1}}

\usepackage{mathrsfs}
\usepackage{makecell}
\usepackage{textcomp}

\def\BibTeX{{\rm B\kern-.05em{\sc i\kern-.025em b}\kern-.08em
    T\kern-.1667em\lower.7ex\hbox{E}\kern-.125emX}}
\makeatletter
\renewcommand\subsection{\@startsection{subsection}{2}{\z@}%
  {0.8ex \@plus 0.5ex \@minus 0.2ex}%
  {0.5ex \@plus 0.2ex}%
  {\normalfont\normalsize\bfseries}}
\makeatother
\begin{document}
\title{MultiFair: Multimodal Balanced Fairness-Aware Medical Classification with Dual-Level Gradient Modulation}

\author{Md Zubair, Hao Zheng, \textit{Member, IEEE}, Grayson W. Armstrong, Lucy Q. Shen, Gabriela Wilson, Yu Tian, \textit{Member, IEEE}, Xingquan Zhu, \textit{Fellow, IEEE},  Min Shi, \textit{Member, IEEE}
\thanks{This work was \rrevise{partially} supported by NSF EPSCoR RII-2531324, IIS-2302786, Louisiana BoRSF-RCS LEQSF(2025-28)-RD-A-19, and LEQSF(2025-28)-RD-A-21.
Md Zubair, Hao Zheng, and Min Shi are with the School of Computing and Informatics, University of Louisiana at Lafayette, Lafayette, LA, USA (e-mail: md.zubair1@louisiana.edu; hao.zheng@louisiana.edu; min.shi@louisiana.edu). 
Gabriela Wilson is with the Louisiana Center for Health Innovation and College of Nursing \& Health Sciences, University of Louisiana at Lafayette, Lafayette, LA, USA (e-mail: gabriela.wilson@louisiana.edu). Grayson W. Armstrong and Lucy Q. Shen are with the Massachusetts Eye and Ear, Harvard Medical School, Boston, MA, USA (e-mail: grayson\_armstrong@meei.harvard.edu, lucy\_shen@meei.harvard.edu). 
Yu Tian is with the Department of Computer Science, University of Central Florida, Florida, USA (e-mail: yu.tian2@ucf.edu).
Xingquan Zhu is with the Department of Electrical Engineering and Computer Science, Florida Atlantic University, Florida, USA (e-mail: xzhu3@fau.edu).
}
\thanks{Corresponding author: Min Shi (e-mail: min.shi@louisiana.edu).
}
}

\maketitle
\newcommand{\modelname}{MultiFair}
\begin{abstract}
Medical decision systems increasingly rely on data from multiple sources to ensure reliable and unbiased diagnosis. However, existing multimodal learning models fail to achieve this goal because they often \revision{overlook} two critical challenges. First, various data modalities may learn unevenly, thereby converging to a model biased towards certain modalities. Second, the model may emphasize learning on certain demographic groups causing unfair performances. The two aspects can influence each other, as different data modalities may favor respective groups during optimization, leading to both imbalanced and unfair multimodal learning. This paper proposes a novel approach called \textit{\modelname} for multimodal medical classification, which addresses these challenges with a dual-level gradient modulation process. \text{\modelname} dynamically modulates training gradients regarding the optimization direction and magnitude at both data modality and group levels. \myrevise{We evaluate MultiFair on three real-world medical classification datasets with diverse demographic attributes,including multiclass classification and missing-modality settings. Experimental results demonstrate its effectiveness.} 

\end{abstract}


\vspace{-3mm}
\section{Introduction}
\label{sec:introduction}
\begin{figure}
  \centering
    \includegraphics[width=0.47\textwidth]{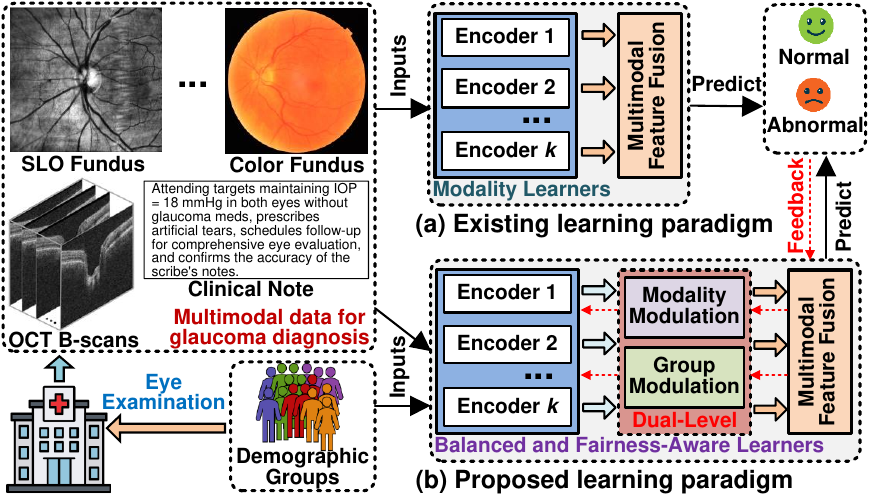}
  \caption{\textbf{The differences between existing and proposed multimodal learning paradigms}. SLO: scanning laser ophthalmoscopy. OCT: optical coherence tomography.}
  \label{fig1}
  \vspace{-4mm}
\end{figure}

Modern medical diagnosis often collects multimodal clinical data to provide a comprehensive assessment of a patient's condition \cite{artsi2024advancing}. Different data modalities, such as genomics, images, textual reports, and physiological signals, can present shared and/or complementary disease biomarkers, which are critical in precision medicine, especially for diagnosing multifactorial diseases \cite{kline2022multimodal}. For example, it is common that an ophthalmologist combines information from multiple observations like retinal fundus photos, optical coherence tomography (OCT) scans, and clinical notes for evaluating glaucoma patients (Fig. \ref{fig1}). OCT detects precise retinal nerve fiber layer thinning, while fundus photo reveals optic disc hemorrhages that OCT may miss \cite{budenz2006detection}, and clinical notes could further add patient-specific information
like family history and symptoms. Integrating them is beneficial to exclude confounding factors such as age and diabetic retinoscopy for reliable diagnostic outcomes\cite{gangwani2016detection}.

To date, numerous works have been conducted to advance multimodal learning \cite{yuan2025survey} and support multimodal medical decision systems \cite{kline2022multimodal, acosta2022multimodal}. The majority of existing works address non-medical tasks with special focus on (1) cross-modal alignment learning, such as CLIP \cite{radford2021learning} and AlignMamba \cite{li2025alignmamba}, and (2) multimodal feature fusion, such as early and late fusion-based approaches \cite{zhao2024deep}. These works essentially take advantage of aligned or enriched information from multiple modalities to train unbiased and robust models. This can overcome limitations of unimodal models that often struggle with noise and incomplete information in individual modalities. Despite remarkable results in scenarios like vision-language models and recommender systems \cite{li2024multimodal}, current methods can be less effective or unreliable in high-stakes biomedical applications for two critical considerations:
\begin{itemize}[leftmargin=1em]
    \item \textbf{Modality Learning Bias:} Different clinical data modalities contain biomarker features that may have uneven contributions to the diagnosis. Unlike physicians who are experienced in integrating useful information across all modalities for a reliable diagnostic outcome, existing gradient-based optimization models are greedy to reduce the global training loss, which may be governed by the dominant modalities in a local minimum. This is evidenced by recent studies \cite{wei2024fly, guo2024classifier}, which indicate that a simple combination of multiple modalities is not advantageous or even worse compared to individual modalities. 
    \item \textbf{Demographic Learning Bias:} AI models raise significant fairness concerns in medical applications, where the learning may be biased to certain demographic groups (\textit{e.g.,} gender or racial groups), while compromising the performance of other groups \cite{luo2024fairclip, jin2024fairmedfm, afolabi2025equity}. This issue could be exacerbated in multimodal learning, as different modalities may unevenly \revision{introduce} unfairness across various groups, making it difficult for the multimodal model to achieve unified fairness optimization.
\end{itemize}

In this paper, we propose to address the above modality and demographic group biases simultaneously. We focus on the medical classification, which is pivotal in many medical decision systems \cite{patricio2023explainable}, while our work can be adapted to other medical tasks. To this end, we propose a novel \textbf{Multi}modal balanced \textbf{Fair}ness-aware \revision{framework} (\textbf{\modelname}) with a dual-level gradient modulation. Fig. \ref{fig1} illustrates the differences between \text{\modelname} and existing multimodal learning models. Several recent works propose balanced multimodal learning models, such as OGM \cite{wei2024fly}, CGGM \cite{guo2024classifier}, and denoising-and-relearning-based \cite{wei2024diagnosing}, which all aim to control different modalities for balanced learning. Our problem and approach are \revision{essentially} different from these works. To the best of our knowledge, this is the first work to jointly address modality and group biases for multimodal medical classification. However, it is non-trivial to address them at the same time, as the two aspects are entangled to impact each other by that (1) imbalanced learning of various modalities may intensify the group bias, and (2) different groups may favor different data modalities for prediction, which in turn reinforces modality bias. To address these challenges, we propose a dual-level gradient modulation mechanism, which mitigates both modality and group bias in a unified framework.

Our major contributions are summarized as follows:
\begin{itemize}[leftmargin=1em]
    \item We propose \text{\modelname}, a novel \revision{framework} incorporating a dual-level gradient modulation process that jointly modulates the training gradients at modality and group levels to optimize model performance and fairness.
    \item We theoretically justify that our framework balances the convergence of the modalities while ensuring fairness across subgroups. 
    
    \item We further verify the effectiveness of the MultiFair through extensive empirical experiments on \rrevise{three} real-world multimodal medical datasets. 
\end{itemize}

\vspace{-2mm}
\section{Related Work}
\textbf{Multimodal Learning.} 
Multimodal learning that integrates diverse modalities such as medical imaging, clinical text, and electronic health records enable\revision{s} more reliable medical predictions. Early works established three foundational paradigms for multimodal fusion: (1) early fusion~\cite{barnum2020benefits} directly combines raw inputs; (2) intermediate fusion~\cite{joze2020mmtm} merges intermediate feature representations; and (3) late fusion~\cite{tsai2019multimodal} integrates independent modality-specific models at the decision level. Building on these strategies, large-scale pretrained vision–language models (VLMs) such as CLIP~\cite{radford2021learning}, ALIGN~\cite{jia2021scaling}, and ViLBERT~\cite{lu2019vilbert} align modalities in shared embedding spaces using dual-stream architectures. Unified transformer-based models, including UNITER~\cite{chen2020uniter} and VisualBERT~\cite{li2019visualbert}, instead perform joint pretraining with a single backbone. Recent advancements such as BLIP-2~\cite{li2023blip} and Flamingo~\cite{alayrac2022flamingo} have further introduced instruction tuning and few-shot reasoning capabilities, while AlignMamba~\cite{li2025alignmamba} and VLMT~\cite{lim2025vlmt} demonstrate state-of-the-art reasoning through scalable cross-modal token fusion. In medical AI, transformers are increasingly applied to imaging and multimodal data. CrossViT~\cite{chen2021crossvit} employs a dual-branch design to fuse small- and large-patch tokens via cross-attention, while MultiViT~\cite{MultiViT} integrates structural MRI and functional connectivity for schizophrenia prediction. Despite their effectiveness in representation learning and disease prediction, these models overlook fairness across demographic subgroups, which is an essential concern in clinical deployment.

\textbf{Multimodal Balanced Learning.} 
In multimodal imbalance learning, faster-learning modalities dominate optimization, while slower ones remain under-optimized. This issue was first highlighted by Wang et al.~\cite{Wang_2020_CVPR}, who proposed Gradient Blending to equalize learning signals across modalities. Subsequent works advanced gradient-level strategies, including On-the-Fly Gradient Modulation (OGM)~\cite{peng2022balanced}, Adaptive Gradient Modulation~\cite{Li_2023_ICCV}, and Classifier-Guided Gradient Modulation (CGGM)~\cite{guo2024classifier}, which jointly calibrate gradient magnitudes and directions. Beyond gradient-based methods, representation and fusion-level strategies (e.g., Online Logit Modulation, Predictive Dynamic Fusion) address modality imbalance at embedding or decision stages. In medical AI, such imbalance degrades performance in tasks like CT–MRI segmentation and Alzheimer’s detection, where domain-adaptive fusion schemes such as Mind the Gap~\cite{su2023mind} and IMBALMED~\cite{francesconi2025class} have been proposed. While these advances enhance performance stability, they are predominantly agnostic to fairness considerations across sensitive attributes (e.g., race and gender), thereby limiting their suitability for equitable healthcare deployment.

\textbf{Fairness-Aware Multimodal Learning.} 
Medical AI raises considerable fairness concerns, as biased predictions can lead to unequal treatment recommendations and amplify existing healthcare disparities. Traditional fairness studies in unimodal medical imaging have exposed systematic demographic biases. For instance, Harvard FairVision~\cite{luo2024fairvision} provides the first large-scale 2D/3D ophthalmic fairness dataset, revealing substantial disparities across race, gender, and ethnicity, and proposes FIN to improve both accuracy and equity. Extending to multimodal settings, Harvard-FairVLMed~\cite{luo2024fairclip} enables fairness analysis of vision-language models and introduces FairCLIP, with an optimal-transport-based approach to balance performance and fairness across demographic groups. Beyond these, domain-general works have also emerged: Kim et al.~\cite{kim2023fairness} propose a fairness regularizer for video-text-audio interviews, Cheong et al.~\cite{cheong2024fairrefuse} introduce FairReFuse for depression detection, and Wu et al.~\cite{wu2024fmbench} develop FMBench to benchmark fairness in MLLMs. In clinical prediction, Wang et al.~\cite{wang2024fairehr} propose FairEHR-CLP to align patient representations with contrastive learning. While these efforts expose critical biases and introduce post-hoc remedies, fairness is often treated as secondary and not embedded within the core fusion process. These gaps motivate us to propose a unified framework that jointly addresses modality imbalance and fairness.

\rrevise{\noindent\textbf{Deficiencies of Existing Works.} Existing multimodal learning approaches primarily focus on effective feature fusion across modalities but do not explicitly address optimization imbalance or fairness issues. Standard multimodal fusion methods rely on early, late, or intermediate fusion strategies and implicitly assume balanced learning across modalities~\cite{barnum2020benefits,joze2020mmtm,tsai2019multimodal, radford2021learning, jia2021scaling, lu2019vilbert,chen2020uniter, li2019visualbert, li2023blip, alayrac2022flamingo, li2025alignmamba, lim2025vlmt, chen2021crossvit,  MultiViT}. Balanced multimodal learning approaches address modality dominance during optimization through both gradient-level and fusion-level strategies \cite{peng2022balanced, Li_2023_ICCV, francesconi2025class, su2023mind}. While effective in stabilizing multimodal training, these methods focus exclusively on modality imbalance and do not account for demographic fairness across sensitive attributes. Conversely, fairness-aware multimodal models, including FairVision~\cite{luo2024fairvision} and FairCLIP~\cite{luo2024fairclip}, incorporate fairness constraints to mitigate group bias, but they do not address modality-level learning imbalance and typically treat fairness independently of multimodal optimization dynamics.} \rrevise{\noindent\textbf{Our Differences.} In contrast, MultiFair differs fundamentally from these prior works by jointly addressing modality imbalance and demographic bias within a unified optimization framework. Specifically, MultiFair introduces a dual-level gradient modulation mechanism that simultaneously modulates gradients across modalities to balance multimodal learning and across demographic groups to enforce fairness. This joint modulation explicitly accounts for modality–group interactions during training, enabling fairness to be embedded directly into the multimodal optimization process rather than applied as a separate or post-hoc constraint. To the best of our knowledge, MultiFair is the first framework to integrate modality balancing and group fairness through coordinated gradient modulation, providing a principled solution for fair and balanced multimodal medical classification.}
\vspace{-2mm}
\section{Problem Definition and preliminaries}

\textbf{Problem Formulation:}
\revision{A multimodal dataset is denoted by}, $\mathcal{\textbf{D}} = \left\{ \left( \textbf{X}_1^{(i)}, \textbf{X}_2^{(i)}, \dots, \textbf{X}_M^{(i)}, g^{(i)}, y^{(i)} \right) \right\}_{i=1}^{N}$,
where $N$ denotes the number of samples, each with $M$ modalities (e.g. images, texts). $\textbf{X}_m^{(i)}$ denotes the $i^{th}$ sample of \( m^{th} \) modality, $g^i$ is the associated demographic subgroup (e.g. gender and race), and ${y}^i$ is the disease detection label. Each modality sample \( \textbf{X}_m^{(i)} \) is processed by its corresponding encoder \( f_m \) to generate a feature representation $h_m^{(i)} = f_m(\textbf{X}_m^{(i)})$. We focus on the medical classification task by predicting disease label $y^{(i)}$ from the fused features $\left\{ h_1^{(i)}, h_2^{(i)}, \dots, h_M^{(i)} \right\} $ of input modalities.

\textbf{Dual-Level Gradient Modulation:} Our work aims to achieve fair and balanced multimodal learning through a dual-level gradient modulation process, represented as $\hat{h}_m^{(i)} = h_m^{(i)} + \Delta_m^{\text{cls},(i)} + \Delta_m^{\text{fair},(i)}$. It integrates classifier-guided modality modulation \( \Delta_m^{\text{cls},(i)} \) and fairness-aware modulation \( \Delta_m^{\text{fair},(i)} \) to ensure modality learning balance and group fairness. The modulated encoder outputs are integrated using a multi-head attention mechanism to generate the final prediction. During the training process, the parameters of the fusion model ($\theta^{\mathcal{F}}$) and the specific modality encoder ($\theta^{\phi_m}$) are optimized simultaneously with gradient descent by:
\begin{equation}
\theta^{\mathcal{F}}_{t+1} = \theta^{\mathcal{F}}_t - \alpha \nabla_{\theta^{\mathcal{F}}} \mathcal{L}(\theta^{\mathcal{F}}_t)
\label{fusion_update}
\end{equation}
\begin{equation}
\theta^{\phi_m}_{t+1} = \theta^{\phi_m}_t - \alpha \nabla_{\theta^{\phi_m}} \mathcal{L}(\theta^{\phi_m}_t)
\label{update}
\end{equation}
where $\alpha$ is the learning rate of $t^{th}$ iteration, and $\mathcal{L}$ is the loss function. \rrevise{Although a single loss function ($\mathcal{L}$) is used for optimization, gradients for the fusion model and each modality-specific encoder are computed separately.}
\vspace{-3.5mm}
\section{Methodology}
\label{multifair_method}
\begin{figure*}[h]
    \centering
    \includegraphics[width=0.9\linewidth]{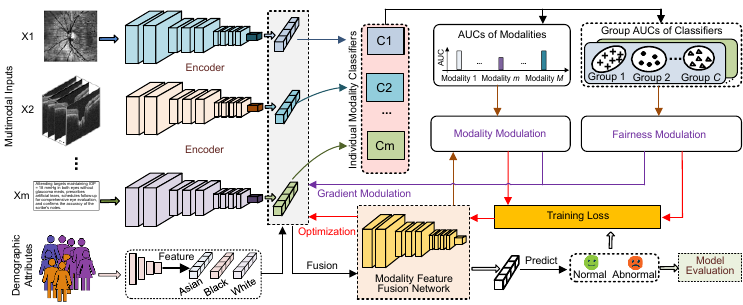}
    \caption{
        \textbf{The proposed MultiFair \revision{framework}}. $X_1, X_2, \ldots, X_m$ represent the modalities. The features of individual encoders are fused by a multi-head attention fusion model for medical classification. Modality-specific classifiers' ($c_1, c_2, \ldots, c_m$) \rrevise{gradient direction, magnitudes, and modality-based AUCs \revision{determine} the modality modulation and fairness components}. The task loss is integrated with the fairness gap, and the direction similarity between the fusion model and the classifiers.
    }
    \label{methodology}
    \vspace{-4mm}
\end{figure*}

This section introduces MultiFair (Fig. \ref{methodology}), which comprises three major parts: (1) Multimodal Medical Classification; (2) Modality Modulation; and (3) Group Fairness Modulation. 
\begin{itemize}[leftmargin=1em]
    \item \textbf{Multimodal Medical Classification:} This part uses a multi-head attention mechanism to combine features of modality encoders to perform multimodal medical classification.
    \item \textbf{Modality Modulation:} This part adopts a classifier-guided gradient modulation process \cite{guo2024classifier} to balance different modalities of multimodal learning. 
    \item \textbf{Group Fairness Modulation:} This part guides equitable learning with modality-based fairness-aware modulation.
\end{itemize}

Taken together, the optimization objective consists of three parts: medical classification loss, modality modulation loss, and average group fairness loss. The combined loss is used during backpropagation from the fusion module to individual modality encoders. The gradient update mechanism, defined in Eq. \ref{gradient_update}, incorporates both modality and fairness balancing factors to guide the modulation of individual encoders. 


\subsection{Multimodal Medical Classification}
\text{\modelname} takes different types of modalities (e.g., images, texts) as inputs and uses modality-specific encoders to extract their feature representations. The features extracted by the modality-specific encoders are then integrated using a multi-head attention fusion model (Fig.~\ref{methodology}) for disease prediction. The predicted class probabilities $\hat{y}^{(i)}$ for fused features $z^{(i)}$ can be represented as $\hat{y}^{(i)} = \text{softmax}(W z^{(i)} + b)$, where $W$ and $b$ denote the learnable weight and bias parameters, respectively. The medical classification loss is defined as:
\begin{equation}
\mathcal{L}_{\text{task}} = -\frac{1}{N} \sum_{i=1}^{N} \sum_{c=1}^{C} y^{(i)}_{c} \, \log \hat{y}^{(i)}_{c}
\label{task_loss}
\end{equation}
where $N$ is the sample size, $C$ is the number of classes and $y^{(i)}$ is the true label.  

\subsection{Modality Modulation}
To achieve balanced multimodal learning, \text{\modelname} considers both the magnitude and direction of gradient update. 

\subsubsection{Modality Balancing Factor}
Let $\mathcal{A}_i^t$ denote the AUC for $i$-th modality at $t^{th}$ iteration.  The learning speed, measured by the change in $AUC(\Delta \mathcal{A})$, can be represented as:
\begin{equation}
\begin{aligned}
\Delta \mathcal{A}_i^{t+1} &= \mathcal{A}_i^{t+1} - \mathcal{A}_i^{t}
\end{aligned}
\label{magnitude}
\end{equation}
where $\Delta A_i^t$ denotes the change in AUC for modality $i$ at iteration $t$.
Based on the \myrevise{magnitude of AUC variation ($| \Delta \mathcal{A}_k^t|$)}, we introduce a balancing factor $\mathcal{B}_i^t$, which assigns higher weights to modalities with lower AUC variation to modulate their gradient updates:
\begin{equation}
\mathcal{B}_i^t = \rho \cdot \frac{\sum_{k=1,\, k \neq i}^{M} | \Delta \mathcal{A}_k^t|}{\sum_{k=1}^{M} | \Delta \mathcal{A}_k^t| +\epsilon}
\label{balancing_factor}
\end{equation}
where $\rho$ is a fixed hyperparameter that controls the strength of the balancing mechanism and is defined prior to training, \myrevise{$\epsilon$ is a very small constant (e.g. $1e-8$) to avoid division by zero}, and $M$ represents the total number of modalities.

When modality $i$ \myrevise{has smaller AUC-change magnitude} (i.e., \myrevise{$|\Delta \mathcal{A}_i^t|$} is small), the ratio of the numerator to the denominator in Eq. \ref{balancing_factor} becomes larger because the numerator excludes the $i$-th modality AUC-change magnitude \myrevise{$|\Delta A_i^t|$},
leading to a higher balancing factor $B_i^t$. This, in turn, increases the gradient modulation for the encoder of modality $i$. Conversely, modalities with higher AUC-\myrevise{change magnitude} receive lower balancing factors. We use this balancing factor into Eq. \ref{update} to modulate \revision{the} modality-specific encoder as follows:
\begin{equation}
\theta^{\phi_i}_{t+1} = \theta^{\phi_i}_t - \alpha \mathcal{B}_i^t\nabla_{\theta^{\phi_i}} \mathcal{L}(\theta^{\phi_i}_t).
\label{update_cggm}
\end{equation}

\subsubsection{Gradient Direction Modulation}
Gradient direction modulation ensures alignment between gradients of modality encoders and the fusion model. Modulation loss $\mathcal{L}_{gm}^{(t)}$ measures the difference between their gradient alignment by: 
\begin{equation}
\mathcal{L}_{gm}^{(t)} = \frac{1}{M} \sum\nolimits_{{i=1}}^{{M}} \left\{ \mathcal{B}_{i}^{(t)} - \mathcal{B}_{i}^{(t)} \cdot \text{sim} \left( \nabla_{\theta_{\mathcal{F}}^{(t)}} \mathcal{L},\ \nabla_{\theta_{f_i}^{(t)}} \mathcal{L} \right) \right\}
\label{direction}
\end{equation}
where $\nabla_{\theta_{f_i}^{(t)}}\mathcal{L}$  and $\nabla_{\theta_{\mathcal{F}}^{(t)}}\mathcal{L}$ are gradients of parameters of the modality encoder $i$ and the fusion module, respectively. The function $sim(.,.)\in[-1,1]$ calculates the cosine similarity between the two gradients. Higher alignment between the gradients of modality encoders leads to a lower modulation loss, whereas misalignment results in a greater penalty. This loss term is incorporated into the fusion loss ($\mathcal{L}_{task}$) with a trade-off hyperparameter $\lambda_{gm}$ by:
\begin{equation}
\mathcal{L}^{(t)} = \mathcal{L}_{\text{task}}^{(t)} + \lambda_{gm} \cdot \mathcal{L}_{gm}^{(t)} .
\label{total_loss}
\end{equation}

\vspace{-2mm}
\subsection{Group Fairness Modulation}
We ensure group fairness by adding a modality-specific modulation factor and a fairness loss term to Eq.~\ref{total_loss}. 

\subsubsection{Fairness-Balancing Factor}
\label{fairness_balancing}
We balance the AUC scores across different groups during training to promote group fairness. To this end, we adopt a differentiable surrogate AUC rather than the traditional non-differentiable AUC \cite{yuan2021large}.

\rrevise{
\noindent\textbf{Surrogate AUC:}
Since AUC is non-differentiable, we estimate group and modality-wise performance using a differentiable pairwise ranking surrogate AUC computed at the mini-batch level \cite{yuan2021large}. For modality encoder $i$ and demographic group $g$, let $\mathcal{P}_{g,i}^{(t)}$ and $\mathcal{N}_{g,i}^{(t)}$ denote the sets of positive and negative samples in the current mini-batch at iteration $t$, with predicted scores $s_{p,i}$ and $s_{n,i}$, respectively. The batch-level surrogate AUC is defined as: 
{\small
\begin{equation}
\mathrm{AUC}^{\text{batch, (t)}}_{g,i}
= 1 - \frac{1}{|\mathcal{P}_{g,i}^{(t)}||\mathcal{N}_{g,i}^{(t)}|}
\sum_{p} \sum_{n}
\log\!\left(
1 + e^{-\frac{s_{p,i}-s_{n,i}}{T}}
\right)
\end{equation}
}
\noindent where $T>0$ is a temperature parameter controlling the smoothness of the approximation (set to $T=1$ in all experiments). This formulation replaces the traditional AUC with a smooth logistic loss, yielding a threshold-free, differentiable ranking measure. If a mini-batch contains only one class, the surrogate AUC is set to $0.5$ for stability. Pseudocode is given in the Appendix \ref{appendix_algo} (Algorithm \ref{alg:surrogate_auc})}.

\textbf{{Exponential Moving Average of AUC:}}  
For every group $g$ and modality $i$, we monitor the performance of the encoders using an exponential moving average (EMA) of a surrogate AUC score. 
The group-based EMA AUC is represented as:  
\begin{equation}
\text{AUC}_{g_i}^{\text{EMA}, (t+1)} = s \cdot \text{AUC}_{g_i}^{\text{EMA}, (t)} + (1 - s) \cdot \text{AUC}_{g_i}^{\text{batch}, (t+1)}
\label{ema}
\end{equation}
where $s \in [0,1)$ is the smoothing factor at iteration $t$. The EMA of AUC reduces abrupt fluctuations by blending the previous EMA value with the current batch AUC. A larger $s$ gives more weight to past performance, whereas a smaller $s$ places greater emphasis on the current batch AUC for group $g$.
It can provide a stable and continuous estimate of model performance across groups \cite{polyak1992acceleration}. \rrevise{Further, the stability of the EMA-based AUC while training the model has been shown in the section \ref{AUC_stability}.} The average EMA AUC across all demographic groups is defined as: 
\begin{equation}
\overline{\text{AUC}}^{\text{EMA},(t)}_i = \frac{1}{G} \sum\nolimits_{g=1}^{G} \text{AUC}_{g_i}^{\text{EMA}, (t)}
\label{avg_g}
\end{equation}
where $G$ indicates the demographic sub-groups. 

\noindent \textbf{Fairness-Aware Modulation Factor:}  
For modality encoder $i$ and group $g$, the modulation factor is computed by:
\begin{equation}
F_i^{(g)} = 1 + \delta \cdot \frac{ \overline{\text{AUC}}^{\text{EMA}}_i - \text{AUC}_{g_i}^{\text{EMA}} }{ \tau }
\label{fairness_factor}
\end{equation}
where $\delta$ controls the modulation strength, \rrevise{$\tau$ is \revision{a} predefined task-level fairness threshold that represents the maximum allowable group-wise AUC disparity and is applied uniformly across all demographic groups.} \myrevise{Here, we assume $F_i^{(g)}$ is bounded and nonnegative during training by using moderate values of $\delta$ and $\tau$, and clipping if needed.} $F_i^{(g)}$ dynamically adjusts the learning for group-modality pairs. When a group’s AUC is below the modality average, the factor increases to emphasize updating the respective group. For well-performing groups, the factor remains close to 1 with limited influence. This mechanism promotes fairness by prioritizing underperforming groups during training.

\textbf{Group-Proportional Modulation:}  
During training, each batch (batch size $N$) contains a varying number of group samples ($N_g$). Let $p_g=\frac{N_g}{N}$ denote the proportion of samples from group $g$ in the current batch. Next, the aggregated fairness modulation factor for modality $i$ is defined by:
\begin{equation}
f_i^{\text{batch}} = \sum\nolimits_{g=1}^G p_g \cdot F_i^{(g)}.
\label{fair_factor}
\end{equation}

\subsubsection{Overall Fairness Loss}
We define a loss term that captures the overall fairness loss across both modalities and groups. The overall fairness loss factor is given as:
\begin{equation}
\mathcal{F}_G^{(t)} = \frac{1}{M} \sum_{i=1}^{M} \left( \frac{1}{G} \sum_{g=1}^{G} \left| \text{AUC}_{g_i}^{\text{EMA}, (t)} - \overline{\text{AUC}}_{i}^{\text{EMA}, (t)} \right| \right).
\label{fair_loss}
\end{equation}
The overall fairness loss $\mathcal{F}_G^{(t)}$ is calculated as the mean absolute deviation of each group’s $\text{AUC}^{\text{EMA}}$ from the average $\overline{\text{AUC}}^{\text{EMA}}$ across demographic groups for a given modality. Therefore, it captures both group and modality-level fairness. 

\subsection{Training and Optimization}
\label{optimization}
\textbf{MultiFair Loss Function:}
MultiFair optimization combines prediction loss ($\mathcal{L_\text{task}}$), modality modulation loss ($\mathcal{L}_\text{gm}$), and \rrevise{fairness} loss ($F_G$). The total loss is given by adding the fairness loss to the Eq. \ref{total_loss} as follows:
\begin{equation}
\mathcal{L}_{\text{total}}^{(t)} = \mathcal{L}_{\text{task}}^{(t)} + \lambda_{gm} \cdot \mathcal{L}_{gm}^{(t)} + \lambda_f \cdot \mathcal{F}_G^{(t)}
\label{final_loss}
\end{equation}
where $\lambda_{gm}$ and $\lambda_f$ are weights to balance the gradient modulation loss and fairness penalty.

\textbf{Gradient Optimization:}
Fairness modulation factor $f^{\text{batch}}$ (Eq.\ref{fair_factor}) is used to scale the encoder gradients as a multiplying factor of the Eq. \ref{update_cggm} and represented as the Eq. \ref{gradient_update}. 
\begin{equation}
\theta_{\phi_i}^{t+1} = \theta_{\phi_i}^{t} - \alpha \, B_i^t \, f_i^{\text{batch}} \, \nabla_{\theta_{\phi_i}} \mathcal{L}(\theta_{\phi_i}^t)
\label{gradient_update}
\end{equation}
where $\mathcal{B}_i^t$ is the modality modulation coefficient, $\alpha$ is the learning rate and $f_i^{batch}$ is the fairness-aware balancing factor. 

To reduce unnecessary computation and redundant parameter updates, fairness modulation is performed selectively. \rrevise{The fairness threshold $\tau$ (introduced in Eq. \ref{fairness_factor} of section \ref{fairness_balancing}) monitors group-wise AUC disparities in the fusion model.} Fairness modulation is triggered only when the difference between best ($Max.AUC_F^g$) and worst ($Min.AUC_F^g$) group AUCs of fusion model exceeds $\tau$  (e.g., $\Delta AUC_F \geq \tau$), such that loss function in Eq.\ref{final_loss} is used and modality encoders are updated as defined in Eq.\ref{gradient_update}. Otherwise, only gradient modulation is used for gradient \rrevise{update} as given in Eq. \ref{update_cggm} and loss function as Eq.\ref{total_loss}. This adaptive strategy enforces fairness optimization only when necessary. Algorithm \ref{algorithm_1} summarizes the training procedure of \text{\modelname}. \rrevise{Additionally, we provide guidelines in the Appendix \ref{multiclass_multimodal} to train MultiFair for Multiclass classification with missing modalities.} We further provide a theoretical justification showing that the stabilized update rule admits bounded descent behavior under deterministic smoothness and boundedness assumptions.

\begin{algorithm}[t]
\label{algorithm_1}
\DontPrintSemicolon
\caption{Training of MultiFair}
\KwIn{Multimodal dataset $\mathcal{D}$ with $N$ samples with modalities $\{X_1, ..., X_M\}$, associated demographic groups, and labels $y$.}
\KwOut{Fair disease classification.}

Initialize parameters: modality encoders $\theta^{\phi_m}$, fusion model $\theta^F$, training epochs $I$, batch size $K$, and fairness gap $\tau$.

\For{$i \in [1,I]$}{
    \For{$j \in [1, N/K]$}{
        $h_m^{(i)} \leftarrow$ extract modality features as  $f_m(\textbf{X}_m^{(i)})$. \;
        $\hat{y}^{(j)} \leftarrow$ prediction of fused modality encoders.\;
        $\mathcal{L}_{\text{task}}^{(t)}\leftarrow$ compute prediction loss by Eq. \ref{task_loss}. \;
        $\mathcal{B}_i^t,\mathcal{L}_{gm}^{(t)}\leftarrow$ calculate modality balancing factor and modulation loss based on Eqs.\ref{balancing_factor} \& \ref{direction}. \;
        $\Delta AUC_F \leftarrow$ ($Max.AUC_F^g - Min.AUC_F^g$) \;
        \eIf{$\Delta \text{AUC}_F \geq \tau$}{
         $f_i^{batch},F_G \leftarrow$ find the fairness modulation factor and fairness loss by Eqs. \ref{fair_factor} \& \ref{fair_loss}. \;
            $\mathcal{L}^{(t)} \leftarrow$ loss function as Eq. \ref{final_loss} \;
            $\theta_{\phi_i}^{t+1} \leftarrow$ update parameters by Eq. \ref{gradient_update}.
            
        }{
            $\mathcal{L}^{(t)} \leftarrow$  loss function as Eq. \ref{total_loss}\;
            $\theta_{\phi_i}^{t+1} \leftarrow$ update parameters by Eq. \ref{update_cggm}\;
        }
        \textbf{End} \;
        Backpropagate to update model parameters \;
    }
    \textbf{End} \;
}
\textbf{End}

\end{algorithm}

\textbf{Theoretical Analysis:}
Let the total loss ($\mathcal{L}_{\text{total}}^{(t)}$) in Eq. \ref{final_loss}, be $L'$-Lipschitz smooth \cite{nesterov2018lectures}, where $\mathcal{L}_{\text{task}}, \mathcal{L}_{gm},$ and $\mathcal{F}_G$ are $L$ -smooth. So,  $L' = L + \lambda_{gm} L + \lambda_f L$. \myrevise{Under the stabilized balancing factor in Eq.\ref{balancing_factor}, the modality coefficient satisfies $0 \le B_i^{(t)} < \rho$, since the numerator is nonnegative and the denominator is strictly positive due to $\epsilon$.} \myrevise{Therefore,} the combined modulation factor satisfies \myrevise{$0 \le \beta^{(t)} = B_i^{(t)} \cdot f_i^{\text{batch}} \leq \beta_{\max}$}, where $\beta_{\max} = \rho \cdot (1 + \delta / \tau)$ (derived from Eq. \ref{balancing_factor}, \ref{fairness_factor} and \ref{gradient_update}).  And fairness modulation is triggered when $\Delta \text{AUC}_F \geq \tau$. We denote the modulation and fairness loss function as $f_j \in \{\mathcal{L}_{gm}, \mathcal{F}_G\}$, and coefficients as
$\lambda_j \in \{\lambda_{gm}, \; \lambda_f\}$. For a learning rate $\alpha < \frac{2}{L' \beta_{\max}}$, MultiFair guarantees the following  theorem: 

\noindent\subsubsection*{Theorem}
\myrevise{ If $\nabla f_j \neq 0$ and $\beta^{(t)} > 0$, then the term
$-\alpha \beta^{(t)} \lambda_j \|\nabla f_j\|^2$ \revision{provides a negative first-order descent contribution towards reducing}
$f_j$. Under the smoothness and boundedness assumptions, $f_j$
admits bounded \revision{descent updates in the deterministic setting}, while $\beta^{(t)} = 0$ corresponds to a null update.}

\vspace{2mm}
\subsubsection*{Proof of the Theorem}
By the Lipschitz continuity of the gradient \cite{nesterov2018lectures}, each $f_j$ is L-smooth, and its gradient ($\theta$) satisfies the following inequality: 
\begin{equation}
\|\nabla f_j(\theta^{(t+1)}) - \nabla f_j(\theta^{(t)})\|
\;\leq\; L \|\theta^{(t+1)} - \theta^{(t)}\|.
\label{lipschitz}
\end{equation}

\noindent As $f_{j}$ satisfies inequality \ref{lipschitz}, it holds the \textbf{\textit{ descent lemma \cite{nesterov2018lectures}}} and can be described as:
 \begin{align}
f_{j}(\theta^{(t+1)}) \leq\ 
& f_{j}(\theta^{(t)}) 
+ \nabla f_{j}(\theta^{(t)})^\top 
(\theta^{(t+1)} - \theta^{(t)}) \notag \\
&+ \frac{L_2}{2} \| \theta^{(t+1)} - \theta^{(t)} \|^2.
\label{descent_lemma}
\end{align}
From the Eq. \ref{gradient_update}, let $\beta^{(t)}=\, B_i^t \, f_i^{\text{batch}}$ then the equation can be written as, $\theta^{(t+1)} - \theta^{(t)} =  - \alpha \beta^{(t)} \nabla \mathcal{L}_{\text{total}}(\theta^{(t)})$. Substituting Eq. \ref{descent_lemma} with the value, we get the inequality:  
 \begin{align}
f_j(\theta^{(t+1)}) 
\leq\ & f_j(\theta^{(t)}) 
- \alpha \beta^{(t)} 
\nabla f_j^\top 
\nabla \mathcal{L}_{\mathrm{total}} \notag \\
&+ \frac{\alpha^2 (\beta^{(t)})^2 L_2}{2} 
\|\nabla \mathcal{L}_{\mathrm{total}}\|^2.
\label{inequality}
\end{align} 
From the total loss definition in Eq. \ref{final_loss}, its gradient can be represented as:
\begin{equation}
\nabla \mathcal{L}_{\mathrm{total}}
= \nabla \mathcal{L}_{\mathrm{task}}
+ \lambda_{\mathrm{gm}} \nabla \mathcal{L}_{\mathrm{gm}}
+ \lambda_f \nabla F_G.
\label{gradient}
\end{equation}
Hence, the inner product of Eq. \ref{gradient} with respect to $\nabla f_j$ can be represented as:
\begin{equation}
\begin{aligned}
\nabla f_j^\top \nabla \mathcal{L}_{\mathrm{total}}
&= \nabla f_j^\top \nabla \mathcal{L}_{\mathrm{task}}
+ \lambda_{\mathrm{gm}} \nabla f_j^\top \nabla \mathcal{L}_{\mathrm{gm}} \\
&+ \lambda_f \nabla f_j^\top \nabla F_G. 
\label{iner_product}
\end{aligned}
\end{equation}
\revision{From Eq. \ref{iner_product}, the inner product can be decomposed as: $\nabla f_j^\top \nabla \mathcal{L}_{\mathrm{total}}
= \lambda_j\|\nabla f_j\|^2 + R_j$, where $
R_j
=
\nabla f_j^{\top}\nabla \mathcal{L}_{task}
+
\sum_{k \neq j}
\lambda_k \nabla f_j^{\top}\nabla f_k$,} represents the cross-terms that denote the interactions between $\nabla f_j$ and the gradients of the other loss components.

By the \textit{Cauchy-Schwarz} \cite{goodfellow2016deep} inequality and assuming bounded gradients, these cross-terms are bounded by a constant $C>0$ \revision{such that $|R_j| \le C$ ($-C \le R_j \le C$). Considering the lower bound, $R_j \ge -C$}, we obtain:
\begin{equation}
\nabla f_j^\top \nabla \mathcal{L}_{\mathrm{total}}
\geq \lambda_j\|\nabla f_j\|^2 - C.
\end{equation}

\noindent Substituting into the inequality (\ref{inequality}), we obtain the following inequality:
\begin{align}
f_j(\theta^{(t+1)}) 
\leq\ & f_j(\theta^{(t)}) 
- \alpha \beta^{(t)} \big( \lambda_j\|\nabla f_j\|^2 - C \big) \notag \\
&+ \frac{\alpha^2 (\beta^{(t)})^2 L_2}{2} \|\nabla \mathcal{L}_{\mathrm{total}}\|^2.
\label{eqn23}
\end{align}

\noindent Since $\beta^{(t)}$ may vary at each iteration depending on batch composition and demographic scaling, we use its boundedness,
$0 \leq \beta^{(t)} \leq \beta_{\max}$. Using this bound, we \revision{represent inequality (\ref{eqn23}) as:
\begin{equation}
\begin{aligned}
f_j(\theta^{t+1})
\le\;&
f_j(\theta^{t})
-
\alpha \beta^{(t)} \lambda_j \|\nabla f_j\|^2 \\
&
+
\alpha \beta_{\max} C
+
\frac{\alpha^2 \beta_{\max}^2 {L_2}}{2}
\|\nabla L_{\mathrm{total}}\|^2 .
\label{aligned}
\end{aligned}
\end{equation}}

\noindent \revision{Inequality (\ref{aligned}) can be further decomposed as:}  
\begin{equation}
f_j(\theta^{(t+1)}) 
\leq f_j(\theta^{(t)}) 
- \alpha \beta^{(t)}\lambda_j \|\nabla f_j\|^2 + O(\alpha^2)
\end{equation}
which gives a descent step when $\beta^{(t)}>0$ and the
first-order descent component $\lambda_j\|\nabla f_j\|^2$ dominates the
bounded cross terms and the higher-order $O(\alpha^2)$ term.
When $\beta^{(t)} = 0$, the update reduces to a null step, corresponding to a locally stable configuration.

\myrevise{\textbf{Remarks:} Let $f_1=\mathcal{L}_{gm}$ (modality balancing loss) and
$f_2=\mathcal{F}_G$ (fairness loss). The proof is derived for the stabilized
update rule in the deterministic (full-batch) setting under smoothness and
boundedness assumptions. In particular, the balancing factor in
Eq.~\ref{balancing_factor} uses absolute AUC changes and denominator smoothing with
$\epsilon$, ensuring that the modulation term remains finite and
nonnegative. Under these assumptions, each loss term admits bounded
descent updates. Although the total loss
combines modality-level and subgroup-level modulation, we analyze them
separately because each component targets a different source of imbalance.
This separation clarifies how each term reduces its corresponding
disparity under standard optimization assumptions. 
\revision{The theoretical analysis characterizes the stabilized update rule under deterministic assumptions and provides insight into the bounded descent behavior of the proposed modulation. Since MultiFair is trained with stochastic mini-batches and EMA-smoothed fairness estimates, the theoretical result is stated only in terms of bounded descent updates under deterministic assumptions. The empirical results in Appendix \ref{dual_modulation} further illustrate stable training behavior. } 
}

\vspace{-4mm}
\section{Experimental Analysis}
\label{experiment}
\vspace{-1mm}
\begin{table*}[!h]
\centering
\caption{Results on the FairVision dataset. Statistical significance compared to the underlined competitive baseline is indicated by * ($p < 0.05$) and ** ($p < 0.001$).}
\label{fairvision_result}

\footnotesize
\begin{adjustbox}{max width=\textwidth}
{\color{black}
\setlength{\tabcolsep}{2pt}
\begin{tabular}{@{}l l cccccc ccccccc@{}}
\toprule
\multirow{2}{*}{\textbf{Modality}} & \multirow{2}{*}{\textbf{Model}} &
\multicolumn{6}{c}{\textbf{Gender}} &
\multicolumn{7}{c}{\textbf{Race}} \\
\cmidrule(lr){3-8} \cmidrule(lr){9-15}
& & \textbf{AUC}$\uparrow$ & \makecell{\textbf{ES-}\\\textbf{AUC}$\uparrow$} &
\makecell{\textbf{Male}\\\textbf{AUC}$\uparrow$} & \makecell{\textbf{Female}\\\textbf{AUC}$\uparrow$} &
\textbf{DPD}$\downarrow$ & \textbf{DeOdds}$\downarrow$ &
\textbf{AUC}$\uparrow$ & \makecell{\textbf{ES-}\\\textbf{AUC}$\uparrow$} &
\makecell{\textbf{Asian}\\\textbf{AUC}$\uparrow$} & \makecell{\textbf{Black}\\\textbf{AUC}$\uparrow$} & \makecell{\textbf{White}\\\textbf{AUC}$\uparrow$} &
\textbf{DPD}$\downarrow$ & \textbf{DeOdds}$\downarrow$ \\
\midrule

\multirow{4}{*}{\makecell{2D Unimodal \\(slo fundus)}}
  & EfficientNet & 78.13±1.15 & 77.26±1.52 &78.73±1.01  &77.62±1.33  & 2.86±1.86 & 2.66±2.01 & 78.13±1.15 & 71.45±2.04 & 82.88±1.19 & 73.72±1.81 & 78.01±1.26  & 14.12±2.20 &14.17±2.78  \\
  & ResNet       & 77.71±0.54& 76.95±0.59 &78.01±1.02 &77.44±0.38  & 3.23±1.93 & 3.78±1.12 & 77.71±0.54& 72.11±1.66 & 80.88±2.07 &73.41±1.26  &78.04±0.52 & 8.21±4.05 & 11.40±3.42 \\
  & \textbf{\underline{VGG}}          & 79.51±0.88\sym{**} & 78.54±0.86\sym{**} &79.94±0.89\sym{**}  &79.12±1.15\sym{**}  & 4.77±1.54 & 4.60±2.19 & 79.51±0.88\sym{**} & 76.00±1.44\sym{**} &80.66±0.86\sym{**} & 76.29±1.14\sym{**} &79.82±0.74\sym{**}  & 10.11±2.61\sym{*}& 8.29±4.54 \\
  & ViT          & 76.00±2.87 & 75.54±2.70 &76.01±2.95  &75.73±2.73  & 5.66±1.58 & 5.36±2.31 & 76.00±2.87 & 72.43±1.97 &78.83±3.88  &74.35±2.85  &75.87±2.68  & 6.32±2.65 & 7.50±2.23 \\
\midrule

\multirow{4}{*}{\makecell{3D Unimodal \\ (Oct Bscan)}}
  & EfficientNet & 80.38±1.03 & 77.42±1.21 & 82.55±1.08 & 78.61±1.01 & 5.11±1.30 & 7.13±1.96 & 80.38±1.03 & 75.92±1.06 & 84.22±1.70 & 78.69±0.54 & 80.02±1.13 & 12.48±3.26 & 10.97±4.67 \\
  & \textbf{\underline{ResNet}}       & 83.27±0.73\sym{**} & 80.43±0.38\sym{**} & 85.21±1.24\sym{*} & 81.68±0.40\sym{**} & 3.33±2.05 & 5.65±1.44 & 83.27±0.73\sym{**} & 79.35±0.68\sym{*} & 85.10±1.50\sym{*} & 80.47±1.33\sym{**} & 83.32±0.73\sym{**} & 16.77±3.89\sym{*} & 15.63±6.40 \\
  & VGG          & 82.24±1.06 & 79.14±1.58 & 84.40±0.82 & 80.48±1.39 & 4.59±1.81 & 7.99±2.06 & 82.24±1.06 & 77.22±1.43 & 85.84±2.14 & 79.55±1.98 & 82.13±0.90 & 11.79±1.87 & 9.56±2.07 \\
  & ViT          & 80.88±2.38 & 77.92±2.48 & 82.29±3.99 & 79.87±0.86 & 4.31±0.87 & 6.69±1.21 & 80.88±2.38 & 77.00±2.33 & 83.62±3.71 & 78.87±3.41 & 80.75±2.07 & 12.48±2.63 & 8.83±3.47 \\
\midrule

\multirow{4}{*}{\makecell{Baseline \\ Multimodal}}
  & ViT            & 77.65±0.84 & 76.98±0.78 & 77.84±1.27 & 77.46±0.63 & 5.03±2.63 & 5.22±2.68 & 77.65±0.84 & 73.22±1.76 & 80.88±0.52 & 75.07±1.24 & 77.82±0.99 & 8.66±2.85 & 11.24±3.45 \\
  & EfficientNet   & 81.93±0.57 & 79.48±0.67 & 83.51±1.15 & 80.52±0.57 & 2.71±1.12 & 3.87±1.17 & 81.93±0.57 & 78.25±1.39 & 84.16±0.96 & 79.20±1.04 & 81.81±0.90 & 17.76±1.38 & 14.71±4.24 \\
  & \textbf{\underline{ResNet}}         & 82.64±1.12\sym{**} & 80.40±1.60\sym{*} & 84.20±0.94\sym{**} & 81.60±1.17\sym{**} & 2.45±1.80\sym{*} & 4.73±1.30 & 82.64±1.12\sym{**} & 78.69±1.68\sym{*} & 85.48±1.30\sym{*} & 80.54±2.21\sym{*} & 82.63±1.24\sym{**} & 15.62±2.73 & 15.43±4.77 \\
  & VGG            & 79.54±0.89 & 78.84±1.08 & 79.95±0.91 & 79.14±1.00 & 5.89±0.66 & 5.15±0.85 & 79.54±0.89 & 75.13±1.77 & 80.47±2.03 & 75.76±1.86 & 80.00±0.86 & 8.66±2.99 & 7.75±2.35 \\
\midrule

\multirow{2}{*}{\makecell{Fairness-Aware \\ Baseline}} 
  & FairVision 
  & 79.96±0.62 & 78.82±0.59 & 80.51±0.85 & 79.52±0.90 & 6.65±1.38 & 6.96±1.69
  & 79.69±0.89 & 75.02±1.12 & 82.83±0.39 & 76.72±0.77 & 79.74±1.01 & 9.96±1.12 & 6.96±1.90 \\
  & \textbf{\underline{FairCLIP}}   
  & 84.32±0.79\sym{*} & 81.42±0.67\sym{*} & 86.97±0.36\sym{*} & 83.64±0.52\sym{*} & 3.13±2.67 & 5.10±3.09
  & 84.19±0.63\sym{**} & 79.47±0.69\sym{*} & 87.94±0.94 & 81.85±0.79\sym{**} & 85.22±0.57\sym{*} & 17.76±3.18 & 19.70±6.54 \\
\midrule

\multirow{2}{*}{\makecell{Existing \\ Multimodal}}
  & CrossViT 
  & 83.48±2.43 & 80.59±1.40 & 85.29±3.53 & 81.95±1.62 & 5.52±0.88 & 7.50±1.40 
  & 83.48±2.43 & 77.55±2.28 & 87.73±2.38 & 80.32±2.21 & 83.36±2.20 & 12.51±6.24 & 13.13±3.72 \\
  & \textbf{\underline{MultiViT}} 
  & 84.18±1.06\sym{*} & 81.44±1.32\sym{*} & 86.02±0.99\sym{*} & 82.65±1.21\sym{*} & 4.84±1.01 & 6.69±1.42 
  & 84.18±1.06\sym{*} & 77.90±1.23\sym{**} & 88.76±1.67 & 80.93±0.55\sym{*} & 84.01±1.18\sym{*} & 15.80±2.53 & 17.37±6.09 \\
\midrule

\multirow{3}{*}{\makecell{Balanced \\ Multimodal}}
  & OPM  
  & 84.10±0.72 & 81.52±0.52 & 85.82±0.94 & 82.02±2.11 & 3.74±1.13 & 4.49±1.55
  & 84.10±0.72 & 78.63±1.24 & 87.26±0.95 & 81.09±0.65 & 84.01±0.65 & 15.27±2.81 & 12.92±3.76 \\
  & OGM  
  & 83.89±0.86 & 81.49±0.69 & 85.55±1.18 & 81.84±2.17 & 4.50±0.87 & 5.85±1.41
  & 83.89±0.86 & 78.49±0.60 & 87.10±1.69 & 80.66±0.73 & 83.75±0.98 & 13.41±4.60 & 10.68±6.60 \\
  & \textbf{\underline{CGGM}} 
  & 84.53±1.05\sym{*} & 82.00±0.43\sym{*} & 87.26±1.37 & 83.92±0.67\sym{*} & 4.09±1.18 & 5.60±1.33
  & 84.53±1.05\sym{*} & 79.58±1.17\sym{*} & 89.14±1.05 & 82.35±1.03\sym{*} & 85.20±0.89\sym{*} & 14.45±2.01 & 13.63±5.51 \\
\midrule

\textbf{Proposed} 
  & \textbf{MultiFair} 
  & \textbf{86.07±0.57} & \textbf{83.03±0.25} & \textbf{88.05±0.84} & \textbf{84.39±0.35} & \textbf{4.53±1.22} & \textbf{6.32±1.36}
  & \textbf{86.31±0.36} & \textbf{82.02±0.48} & \textbf{89.04±1.13} & \textbf{83.15±0.72} & \textbf{85.97±0.45} & \textbf{16.38±2.80} & \textbf{14.53±6.66} \\
\bottomrule

\end{tabular}
}
\end{adjustbox}
\vspace{-3mm}
\end{table*}

This section introduces datasets, comparative methods, experimental settings, results, and ablation studies.

\subsection{Datasets}
\label{data_description}
We use three multimodal datasets: 1) FairVision \cite{luo2024fairvision}, 2) FairCLIP\cite{luo2024fairclip}, and \rrevise{3) CheXpert} \cite{rajpurkar2017chexpert}, described as follows:
\begin{itemize}[leftmargin=1em]
    \item \textbf{FairVision \cite{luo2024fairvision}:} The dataset comprises 10,000 paired OCT and SLO fundus samples, each corresponding to a unique patient. Each 3D OCT image consists of 200 B-scans, and each B-scan has a resolution of 200 × 200 pixels. And the 2D SLO fundus images also have a resolution of 200×200 pixels. Demographic distribution based on self-reported information includes 57.0\% female and 43.0\% male patients; racially, 8.5\% identify as Asian, 14.9\% as Black, and 76.6\% as White. Glaucoma labels are assigned based on comprehensive clinical evaluation, and there are 51.3\% non-glaucoma and 48.7\% glaucoma cases. The dataset contains 6000 training and 1000 validation samples. \myrevise{The test set contains 3,000 samples, including 1,716 female and 1,284 male patients, with a racial distribution of 251 Asian, 431 Black, and 2,318 White patients.} We use the standard dataset split for the model training.
    \item \textbf{FairCLIP \cite{luo2024fairclip}:} This dataset contains 10,000 patient records, each consisting of a 2D SLO fundus image (200×200 resolution) and an associated clinical note. Demographic information, based on self-report, indicates a gender distribution of 56.3\% female and 43.7\% male. The racial composition includes 76.9\% White, 14.9\% Black, and 8.2\% Asian patients. Glaucoma diagnoses were determined through clinical evaluation, and  50.5\% of patients identified as having glaucoma and 49.5\% as non-glaucoma.\rrevise{The dataset contains 7000 training and 1000 validation data. \myrevise{The test set contains 2,000 samples, including 1,096 female and 904 male patients, with a racial distribution of 1,537 White, 305 Black, and 158 Asian patients.} We use the standard dataset split for the model training.}

    \item \rrevise{\textbf{CheXpert \cite{rajpurkar2017chexpert}:} \label{Chexpert_dataset}
    We use the CheXpert-v1.0-small dataset, which is a large-scale multi-label chest X-ray dataset with annotations for \myrevise{13} clinical observations. In our setup, each sample is defined at the patient--study level, comprising 187,841 unique samples and 223,648 images. The training set contains 168,834 samples, while the test set includes 18,807 samples. \revision{Since race or ethnicity annotations are not available in the dataset, subgroup analysis is limited to gender demographic groups.} The test set was constructed using 10\% unique patients \myrevise{(13,471 males and 8,908 females)}, ensuring no patient overlap between the train and test sets to prevent data leakage. Approximately 83\% of samples are frontal-only, while around 17\% contain both frontal and lateral views (two modalities), resulting in naturally missing lateral modalities. The dataset is demographically imbalanced, with approximately 59--60\% male and 40--41\% female samples, making it a realistic benchmark for fairness-aware multimodal learning.}
\end{itemize}
\noindent This study complied with the guidelines outlined in the Declaration of Helsinki. In light of the study’s retrospective design, the requirement for informed consent was waived.

\begin{table*}[!h]
\centering
\caption{Results on 2D fundus images and clinical text notes on FairCLIP dataset. Statistical significance compared to the underlined competitive baseline is indicated by * ($p < 0.05$) and ** ($p < 0.001$).}
\label{fairCLIP_dataset}
\footnotesize
\begin{adjustbox}{max width=\textwidth}
\color{black}
\renewcommand{\arraystretch}{0.9}
\begin{tabular}{@{}l l cccccc ccccccc@{}}
\toprule
\multirow{2}{*}{\textbf{Modality}} & \multirow{2}{*}{\textbf{Model}} &
\multicolumn{6}{c}{\textbf{Gender}} &
\multicolumn{7}{c}{\textbf{Race}} \\
\cmidrule(lr){3-8} \cmidrule(lr){9-15}
& & \textbf{AUC}$\uparrow$ & \makecell{\textbf{ES-}\\\textbf{AUC}$\uparrow$} &
\makecell{\textbf{Male}\\\textbf{AUC}$\uparrow$} & \makecell{\textbf{Female}\\\textbf{AUC}$\uparrow$} &
\textbf{DPD}$\downarrow$ & \textbf{DeOdds}$\downarrow$ &
\textbf{AUC}$\uparrow$ & \makecell{\textbf{ES-}\\\textbf{AUC}$\uparrow$} &
\makecell{\textbf{Asian}\\\textbf{AUC}$\uparrow$} & \makecell{\textbf{Black}\\\textbf{AUC}$\uparrow$} & \makecell{\textbf{White}\\\textbf{AUC}$\uparrow$} &
\textbf{DPD}$\downarrow$ & \textbf{DeOdds}$\downarrow$ \\
\midrule
\multirow{4}{*}{\makecell{2D Unimodal \\ (slo fundus)}} 
    & EfficientNet 
    & 79.44±0.28 & 76.22±0.58 & 81.55±0.37 & 77.59±0.39 & 1.04±0.67 & 4.59±0.68
    & 79.44±0.28 & 74.78±1.46 & 81.96±1.80 & 75.73±0.51 & 79.73±0.28 & 9.55±1.24 & 7.10±1.98 \\
    & ResNet       
    & 80.13±1.14 & 75.73±1.29 & 83.20±1.03 & 77.39±1.26 & 1.44±0.70 & 6.60±0.78
    & 80.13±1.14 & 76.83±1.30 & 81.35±2.00 & 77.51±0.71 & 80.23±1.50 & 12.48±4.25 & 10.71±4.45 \\
    & \textbf{\underline{VGG}}          
    & 82.26±0.69\sym{**} & 78.43±1.12\sym{**} & 84.90±0.53\sym{**} & 80.00±0.97\sym{**} & 2.08±0.83 & 6.91±2.29
    & 82.26±0.69\sym{**} & 78.22±1.07\sym{**} & 84.49±1.72\sym{**} & 79.48±0.77\sym{**} & 82.36±0.68\sym{**} & 13.10±3.04 & 10.48±0.81 \\
    & ViT          
    & 81.38±0.72 & 78.13±0.63 & 83.70±0.65 & 79.31±0.82 & 1.79±1.29 & 5.94±1.14
    & 81.38±0.72 & 75.62±1.87 & 83.32±1.34 & 76.42±1.69 & 82.13±0.65 & 9.08±2.60 & 10.99±4.87 \\ 
\midrule
\multirow{2}{*}{\makecell{Text}} 
     & \textbf{\underline{Google-Bert}}  
     & 87.10±0.82\sym{*} & 86.38±0.55\sym{**} & 87.54±1.02\sym{**} & 86.70±0.67\sym{**} & 1.92±0.80 & 3.84±0.99
     & 87.10±0.82\sym{**} & 79.20±1.40\sym{**} & 93.66±0.64\sym{*} & 89.39±0.88\sym{**} & 85.97±0.89\sym{*} & 12.36±1.39 & 15.62±1.88 \\
     & DistilBert   
     & 86.92±0.31 & 86.15±0.24 & 87.42±0.48 & 86.53±0.23 & 0.72±0.53 & 2.14±1.11
     & 86.92±0.31 & 79.14±0.88 & 93.63±0.63 & 88.99±0.52 & 85.85±0.41 & 10.91±1.27 & 15.52±1.78 \\
\midrule

\multirow{4}{*}{\makecell{Baseline \\ Multimodal}}

    & \textbf{\underline{EfficientNet}} 
    & 89.21±0.48\sym{**} & 87.83±0.46\sym{**} & 90.04±0.76\sym{**} & 88.47±0.39\sym{**} & 1.24±0.70 & 4.33±0.66
    & 89.21±0.48\sym{**} & 85.18±1.37\sym{*} & 93.20±1.22\sym{*} & 89.15±0.87\sym{*} & 88.81±0.48\sym{**} & 8.80±2.20 & 6.35±1.99 \\
    & ResNet       
    & 87.98±0.79 & 86.18±1.54 & 89.12±0.45 & 87.02±1.21 & 1.98±0.54 & 4.35±1.47
    & 87.98±0.79 & 84.62±1.41 & 90.63±0.59 & 88.75±0.88 & 87.79±1.01 & 8.39±4.09 & 11.47±3.63 \\
    & VGG          
    & 82.76±0.85 & 79.15±1.22 & 85.19±0.64 & 80.63±1.11 & 0.74±0.46 & 4.94±0.91
    & 82.76±0.85 & 79.35±1.75 & 82.98±1.31 & 80.26±1.56 & 83.11±0.96 & 10.50±3.22 & 9.79±4.48 \\ 
    & ViT          
    & 88.87±1.51 & 87.37±1.10 & 89.80±1.87 & 88.08±1.27 & 1.83±0.97 & 4.24±1.50
    & 88.87±1.51 & 85.42±2.55 & 91.88±1.70 & 88.41±0.75 & 88.79±1.72 & 8.67±4.63 & 9.97±3.30 \\

\midrule

\multirow{2}{*}{\makecell{Fairness-Aware\\ Baseline} }
 & FairCLIP
 & 65.53±1.90 & 62.53±2.40 & 68.20±1.69 & 63.38±2.28 & 2.79±2.21 & 6.11±3.57
 & 65.79±1.91 & 62.88±0.93 & 66.76±4.90 & 64.38±1.10 & 66.01±2.34 & 10.97±5.26 & 15.85±1.86 \\
 & \textbf{\underline{FairVision}}
 & 86.34±2.10\sym{*} & 83.33±2.04\sym{*} & 88.28±2.12\sym{*} & 84.66±2.09\sym{*} & 0.55±0.40 & 4.64±1.04
 & 86.86±1.76\sym{*} & 82.84±2.46\sym{*} & 86.10±0.94\sym{**} & 83.54±2.80\sym{*} & 87.22±1.84\sym{*} & 18.94±4.40\sym{*} & 16.35±4.69 \\

\midrule
\multirow{2}{*}{\makecell{Existing \\ Multimodal}} 
    & CLIP     
    & 75.59±2.23 & 72.71±1.39 & 77.46±3.53 & 73.98±1.15 & 1.46±1.61 & 6.12±2.57
    & 75.59±2.23 & 72.99±2.72 & 73.27±2.38 & 75.60±1.37 & 75.80±2.20 & 9.28±4.69 & 9.23±2.78 \\
    & \textbf{\underline{BLIP-2}}   
    & 83.71±0.42\sym{**} & 81.06±1.20\sym{**} & 85.48±0.27\sym{**} & 82.19±0.87\sym{**} & 1.40±0.86 & 5.63±1.69
    & 83.71±0.42\sym{**} & 81.11±0.90\sym{**} & 83.47±1.53\sym{**} & 81.74±0.54\sym{**} & 83.93±0.47\sym{**} & 10.31±3.08 & 7.51±2.45 \\ 
\midrule

\multirow{3}{*}{\makecell{Balanced  \\ Multimodal}} 
    & OPM          
    & 80.97±0.59 & 77.90±1.15 & 83.08±0.57 & 79.14±0.90 & 3.00±2.05 & 7.33±2.58
    & 80.97±0.59 & 76.07±1.43 & 83.97±1.61 & 77.86±1.25 & 81.29±0.82 & 9.00±5.47 & 11.04±3.37 \\ 
    & \textbf{\underline{OGM}}          
    & 89.47±0.34\sym{**} & 87.34±0.98\sym{*} & 90.78±0.27\sym{**} & 88.34±0.67\sym{*} & 1.73±1.31 & 5.00±2.32
    & 89.47±0.34\sym{**} & 85.69±1.02\sym{*} & 92.34±2.32\sym{*} & 88.38±0.27\sym{**} & 89.30±0.29\sym{**} & 12.11±2.52 & 8.96±1.28 \\ 
    & CGGM         
    & 88.20±3.35 & 86.09±4.69 & 89.57±2.46 & 87.04±4.21 & 2.48±1.14 & 5.67±2.80
    & 88.20±3.35 & 83.38±3.07 & 92.20±5.35 & 86.79±4.78 & 88.13±2.95 & 8.63±3.38 & 9.86±2.59 \\ 

\midrule
\textbf{Proposed} 
    & \textbf{\underline{MultiFair}} 
    & \textbf{91.25±0.44} & \textbf{89.45±0.63} & \textbf{92.37±0.42} & \textbf{90.32±0.55} & \textbf{2.22±1.36} & \textbf{5.39±1.20}
    & \textbf{91.51±0.41} & \textbf{88.36±0.85} & \textbf{94.36±0.67} & \textbf{91.08±0.83} & \textbf{91.33±0.39} & \textbf{11.83±1.91} & \textbf{9.35±2.42} \\

\bottomrule
\end{tabular}
\end{adjustbox}
\vspace{-3mm}
\end{table*}

\subsection{Comparative Methods} We compare MultiFair against a wide range of unimodal, baseline multimodal, fairness-aware, existing multimodal, and balanced multimodal approaches. \myrevise{To ensure a fair comparison, all methods are tuned using a standardized grid search protocol over comparable hyperparameter ranges. 
Model selection is performed based on validation performance, where the best-performing checkpoint is selected for reporting results. The same validation splits, evaluation criteria, and early-stopping protocol are applied consistently for all methods.}

\noindent \textit{\textbf{Unimodal Models}}
\begin{itemize}[leftmargin=1em]
    \item \textbf{EfficientNet \cite{tan2019efficientnet}:} EfficientNet is a deep convolutional neural network designed for scalability, enhancing both accuracy and efficiency by combin\revision{ing} scaling depth, width, and resolution through a compound coefficient.
    \item \textbf{ResNet \cite{he2016resnet}:} ResNet is a deep convolutional neural network that introduces residual connections (skip connections) to enable the training of very deep architectures by mitigating vanishing gradient problems.
    \item \textbf{VGG\cite{simonyan2015vgg}:} VGG is a deep convolutional neural network that uses small convolutional filters to increase network depth and improve performance in image recognition tasks. 
    \item \rrevise{\textbf{Densenet121\cite{huang2017densenet}:} DenseNet121 is a convolutional neural network where each layer receives features from all previous layers, improving feature reuse, gradient flow, and learning efficiency with fewer parameters. Its ability to learn subtle and multi-scale visual features makes it well-suited for medical imaging datasets such as CheXpert \cite{rajpurkar2017chexpert}. In these datasets, many diseases appear visually similar, and the data is often highly imbalanced. This helps the model achieve more robust and generalizable performance.}
  
    \item \rrevise{\textbf{Swin\cite{liu2021swin}}: Swin is a framework that uses a hierarchical, window-based self-attention mechanism to capture both local details and broader contextual information efficiently. This makes it especially appropriate for medical imaging tasks, where disease patterns are often localized, and effective performance depends on capturing context across multiple spatial scales.}
    \item \textbf{ViT\cite{ViT}:} Vision Transformer (ViT) is a transformer-based architecture that treats an image as a sequence of patches, applying self-attention mechanisms to achieve state-of-the-art performance in image recognition.
    \item \textbf{BERT\cite{devlin2019bert}:} Google's BERT classifier is a fine-tuned model that uses the contextualized embedding of the special token, passed through a dense layer with softmax or sigmoid activation, to perform classification tasks.
\end{itemize}

\noindent \textbf{\textit{Baseline Multimodal Models}:} These models are created with fused outputs from separate unimodal encoders (e.g., VGG for images, BERT for text) through simple concatenation. 
 \rrevise{For the FairVision~\cite{luo2024fairvision} dataset, identical image encoders (ViT, VGG, ResNet, and EfficientNet) are used for both OCT and SLO fundus modalities, and the resulting modality-specific features are concatenated to generate the final prediction. In the case of the FairCLIP~\cite{luo2024fairclip} dataset, SLO fundus images are encoded using image encoders (ViT, VGG, ResNet, and EfficientNet), while the corresponding clinical text notes are encoded using a BERT-based text encoder. The image and text features are then concatenated and fed into the classifier to produce the prediction. }

\noindent \textbf{\textit{Fairness-Aware Baseline Multimodal Models}}
\begin{itemize}[leftmargin=1em]
    \item \textbf{FairVision \cite{luo2024fairvision}:} It is a fairness-aware deep learning framework for disease screening that employs fair identity scaling to mitigate demographic bias. \rrevise{In the experiment, we extend FairVision to multimodal settings for both datasets. For the FairVision dataset, OCT and SLO fundus images are encoded using identical image backbones with fair identity scaling applied independently to each modality, and the resulting features are concatenated for classification. Additionally, FairVision is extended to a vision–language setting for the FairCLIP dataset by encoding SLO fundus images with image encoders and associated clinical text notes with a BERT-based text encoder; fair identity scaling is applied at the encoder level, and the image and text embeddings are concatenated to produce the final prediction.}
    \item \textbf{FairCLIP \cite{luo2024fairclip}:} FairCLIP is a fairness-aware vision-language model that extends CLIP by incorporating fairness constraints to reduce demographic bias while maintaining strong multimodal representation learning. \rrevise{It was developed based on the FairCLIP dataset. We also adapted the model to the FairVision dataset. The text encoder is replaced with a ViT-based image encoder. Thus, both OCT and SLO fundus images are encoded using ViT backbones. The resulting modality-specific embeddings are concatenated for prediction, while the fairness constraints and optimization strategy of FairCLIP remain unchanged.}
\end{itemize}

\noindent \textbf{\textit{Existing Multimodal Models without Balanced Learning}}
\begin{itemize}[leftmargin=1em]
    \item \textbf{CrossViT\cite{chen2021crossvit}:} CrossViT is a vision transformer architecture that leverages cross-attention across multi-scale image patches to capture both fine-grained details and global context for improved image recognition.
    \item \textbf{MultiViT \cite{MultiViT}:} It is a vision transformer that fuses different types of images together, combining their strengths to give a more complete and reliable understanding.
    \item \textbf{CLIP\cite{radford2021learning}:} CLIP is a vision–language model that learns shared representations of images and text from large-scale image–caption data. \rrevise{In our experiments, CLIP is initialized with publicly available pretrained weights and is used as a standard vision–language baseline without additional task-specific adaptation.}
    \item \textbf{BLIP-2 \cite{li2023blip}:} BLIP-2 is a vision–language model that connects frozen image encoders with large language models through a lightweight querying transformer, enabling efficient multimodal understanding and generation. \rrevise{We employ BLIP-2 in its standard pretrained configuration without further adaptation to the target task and include it as a representative general-purpose multimodal baseline.}
\end{itemize}

\noindent \textbf{\textit{Existing Multimodal Models with Balanced Learning}}
\begin{itemize}[leftmargin=1em]
    \item \textbf{OPM \cite{wei2025onthefly}:} OPM dynamically modulates both predictions and gradients during training to balance contributions across modalities and improve multimodal learning performance.
    \item \textbf{OGM\cite{peng2022balanced}:} On-the-fly Gradient Modulation (OGM) dynamically adjusts gradients from different modalities during training to prevent dominance of any single modality and achieve more balanced multimodal learning.
    \item \textbf{CGGM \cite{guo2024classifier}:} Classifier Guided Gradient Modulation (CGGM) considers both gradient magnitude and direction modulation to balance multiple modalities.
\end{itemize}

\noindent \textbf{\textit{Robust Missing Modality Handling}}
\begin{itemize}[leftmargin=1em]
    \item \rrevise{\textbf{MedFuse \cite{hayat2022medfuse}:} MedFuse is a multimodal fusion framework designed for healthcare settings where data are often partially paired, and modalities may be missing. It uses modality-specific encoders and performs fusion by treating the available modality embeddings as a sequence, allowing flexible integration without requiring all modalities to be present.  In this work, we adopt MedFuse for the CheXpert \cite{rajpurkar2017chexpert} dataset by modeling the frontal and lateral chest X-ray views as separate modalities for effective fusion under missing-view scenarios.}

    \item \rrevise{\textbf{MoE\cite{shazeer2017moe}:} Mixture of Experts (MoE) is a modular learning framework that consists of multiple specialized sub-networks called experts. A sparse gating mechanism dynamically selects and weights a small subset of these experts for each input. This design enables the model to capture heterogeneous patterns in the data and focus computation on the most relevant experts, making MoE well suited for multimodal and heterogeneous input scenarios, and adaptable to settings with incomplete or variable information.}
\end{itemize}

\subsection{Experimental Settings}
\label{experimental_settings}
\textbf{Dataset and Parameters:} We follow the FairVision \cite{luo2024fairvision} and FairCLIP \cite{luo2024fairclip} paper\revision{s} to split the training, validation, and testing sets. \rrevise{Additionally, we conduct experiments on the multi-level, multi-class CheXpert dataset with missing modalities, using the split described in Section~\ref{data_description}. We perform a compact grid search over a small, predefined set of values for selecting the optimum hyperparameters that optimize validation performance under the fairness objective (Appendix \ref{hyperparameter_selection}). \revision{The EMA smoothing factor (s) is empirically set to 0.9 across datasets to balance stability and responsiveness in the subgroup surrogate-AUC estimates used for fairness modulation.} The MultiFair framework treats the demographic subgroup as the sensitive attribute for fairness optimization; therefore, models are trained and optimized separately for each subgroup to ensure effective fairness optimization without degrading predictive performance.} For the FairVision dataset, we use a projection dimension of 128 with 4 heads in the attention mechanism, and for the FairCLIP dataset, we use 256 projection dimensions with 8 heads. For both datasets, we used the learning rate, $\alpha = 3e^{-5}$, gradient scaling factor, $\rho=1.2$, and gradient modulation, $\lambda_{gm}=0.15$. The fairness threshold ($\tau$), fairness modulation parameter ($\delta$), and fairness penalty ($\lambda_f$) for both gender and race subgroups are set as follows. For FairVision, the hyperparameter values were set to $\tau = 0.04$, \rrevise{$\delta = 0.4$}, and $\lambda_f = 0.5$ for gender, and $\tau = 0.02$, $\delta = 0.6$, and $\lambda_f = 0.6$ for race. For FairCLIP, the corresponding values were \rrevise{$\tau = 0.05$}, $\delta = 0.5$, and $\lambda_f = 0.5$ for both gender and race. \rrevise{In case of the the CheXpert\cite{rajpurkar2017chexpert} dataset, which includes only gender as the demographic attribute, 
we use a learning rate of $\alpha = 3 \times 10^{-5}$, projection dimension of 128 with 4 heads, $\rho = 1.2$, $\lambda_{gm} = 0.15$, $\tau = 0.05$, $\delta = 0.5$, and 
$\lambda_f = 0.3$.} Please refer to our code implementation of MultiFair for detailed parameter settings via GitHub link: \textcolor{blue}{https://github.com/Zubair063/MultiFair}.

\begin{table}[!h]
\centering
\caption{Results on the CheXpert Dataset, and results are macro-averaged over multiple classes. Statistical significance compared to the underlined competitive is indicated by * ($p < 0.05$) and ** ($p < 0.001$). }
\label{chexpert_unimodal_gender}
\footnotesize
\begin{adjustbox}{max width=0.48\textwidth}
\color{black}
\renewcommand{\arraystretch}{0.9}
\begin{tabular}{@{}l l cccccc@{}}
\toprule
\multirow{2}{*}{\textbf{Category}} & \multirow{2}{*}{\textbf{Model}} &
\multicolumn{6}{c}{\textbf{Gender}} \\
\cmidrule(lr){3-8}
& & \textbf{AUC}$\uparrow$ & \makecell{\textbf{ES-}\\\textbf{AUC}$\uparrow$} &
\makecell{\textbf{Male}\\\textbf{AUC}$\uparrow$} & \makecell{\textbf{Female}\\\textbf{AUC}$\uparrow$} &
\textbf{DPD}$\downarrow$ & \textbf{DeOdds}$\downarrow$ \\
\midrule

\multirow{4}{*}{\makecell{Unimodal \\ (\revision{Frontal Chest} \\ \revision{X-Ray})}} 
& \textbf{\underline{Densenet121}}
& 78.90±0.24\sym{**} & 77.93±0.29\sym{*} & 78.69±0.21\sym{**} & 79.17±0.32\sym{*} & 1.01±0.23 & 2.90±0.62 \\
& Efficientnet
& 78.15±0.79 & 77.31±0.73 & 77.91±0.78 & 78.49±0.83 & 0.68±0.17 & 2.10±0.46 \\
& Swin
& 77.82±0.68 & 76.82±0.64 & 77.63±0.66 & 78.09±0.74 & 0.73±0.08 & 1.98±0.22 \\
& ViT
& 76.33±0.69 & 75.56±0.74 & 76.17±0.68 & 76.53±0.69 & 0.69±0.15 & 1.65±0.32 \\
\midrule
\multirow{4}{*}{\makecell{Baseline\\Multimodal}} 
& \textbf{\underline{Densenet121}}
& 78.70±0.52\sym{*} & 77.67±0.55\sym{*} & 78.51±0.52\sym{*} & 78.96±0.53\sym{*} & 1.07±0.20\sym{*} & 2.72±0.45 \\
& Efficientnet
& 78.42±0.31 & 77.71±0.14 & 78.30±0.22 & 78.64±0.41 & 0.89±0.08 & 2.25±1.12 \\
& Swin
& 77.89±0.49 & 76.91±0.55 & 77.65±0.57 & 78.26±0.37 & 0.92±0.15 & 2.52±0.30 \\
& ViT
& 76.62±0.11 & 75.89±0.23 & 76.43±0.18 & 76.85±0.06 & 0.57±0.06 & 1.71±0.61 \\
\midrule

\multirow{2}{*}{\makecell{Fairness Aware \\ Baseline}} 
& \textbf{\underline{FairCLIP}}
& 74.10±1.00\sym{**} & 73.21±1.05\sym{**} & 73.94±0.96\sym{**} & 74.30±1.06\sym{**} & 0.68±0.17 & 1.59±0.49 \\
& FairVision
& 73.17±0.60 & 72.34±0.59 & 73.07±0.61 & 73.31±0.61 & 0.50±0.10 & 0.74±0.19 \\
\midrule
\multirow{3}{*}{\makecell{Balanced\\Multimodal}} 
& \textbf{\underline{OPM}}
& 76.24±0.34\sym{**} & 75.30±0.33\sym{**} & 75.91±0.34\sym{**} & 76.72±0.36\sym{**} & 0.79±0.13 & 1.67±0.26 \\
& OGM
& 68.27±2.10 & 67.37±2.20 & 68.46±2.04 & 67.95±2.18 & 0.53±0.12 & 0.92±0.24 \\
& CGGM
& 75.28±0.63 & 74.39±0.67 & 75.06±0.61 & 75.62±0.63 & 0.73±0.09 & 1.58±0.39 \\
\midrule
\multirow{2}{*}{\makecell{Robust Missing\\Modality}} 
& MedFuse
& 75.04±0.68 & 74.18±0.63 & 74.75±0.67 & 75.43±0.71 & 1.10±0.42 & 2.50±0.42 \\
& \textbf{\underline{MoE}}
& 75.22±0.94\sym{**} & 74.43±0.98\sym{**} & 74.96±0.96\sym{**} & 75.66±0.84\sym{**} & 1.22±0.34 & 2.52±0.47 \\
\midrule
\multirow{1}{*}{\makecell{\textbf{Proposed}}} 
& \textbf{\underline{MultiFair}}
& \textbf{80.73±0.52} & \textbf{79.82±0.57} & \textbf{80.54±0.44} & \textbf{80.93±0.63} & \textbf{0.76±0.13} & \textbf{2.09±0.40} \\
\bottomrule
\end{tabular}
\end{adjustbox}
\vspace{-3mm}
\end{table}

\textbf{Evaluation Metrics.} For predictive performance, we use the Area Under the Receiver Operating Characteristic Curve (AUC), along with the Equity-Scaled AUC (ES-AUC) \cite{luo2024harvard} which adjusts the conventional AUC by incorporating fairness considerations. We also consider other fairness evaluation metrics such as Demographic Parity Difference (DPD) \cite{agarwal2018reductions,agarwal2019fair} and Difference in Equalized Odds (DeOdds) \cite{agarwal2018reductions}, \rrevise{which} quantify disparities across sensitive groups.  \rrevise{For DPD and DeOdds, predicted probabilities are binarized using a single global threshold shared across all demographic groups, which is fixed during evaluation (threshold = 0.5 in the implementation). Group-specific thresholds are not used to avoid artificially manipulating parity.}

\subsection{Experimental Results}
\subsubsection{Overall Comparative performance}
Tables \ref{fairvision_result}, \ref{fairCLIP_dataset}, \rrevise{and \ref{chexpert_unimodal_gender}} report AUC, ES-AUC, and fairness metrics (DPD, DEOdds) for two protected attributes, namely gender and race. Additionally, subgroup AUCs (e.g., gender and race) indicate that MultiFair improves fairness by enhancing the performance of underrepresented groups across the three datasets. All the models have been randomly trained and evaluated for six times and the results are reported as $ Mean \ \pm \ Standard \ Deviation $ to ensure reproducibility. \myrevise{Additionally, paired statistical significance tests were conducted across random experiments using the underlined competitive models for all metrics, including DPD and DeOdds, to verify that the observed improvements are not due to random variation.}   

\rrevise{Our model directly optimizes a surrogate AUC objective, so AUC and ES-AUC are the most appropriate evaluation metrics because they align with the training goal. These metrics assess ranking performance across the full score distribution and are independent of a fixed decision threshold, making them suitable for clinical applications. Since MultiFair does not explicitly optimize fairness at a fixed threshold, improvements in ranking fairness (AUC-based) do not directly guarantee corresponding improvements in DPD or DEOdds.}

\rrevise{Table~\ref{fairvision_result} reports the performance and fairness comparison on the FairVision \cite{luo2024fairvision} dataset over six random experiments, with statistical significance relative to the baseline models based on paired testing. MultiFair achieves the highest overall AUC (86.07\%) and ES-AUC (83.03\%) for gender, as well as the highest overall AUC (86.31\%) and ES-AUC (82.02\%) for race. It also shows the statistically significant improvements in the AUC and ES-AUC over other competitive baseline models. For gender and race subgroups, MultiFair consistently improves balanced ranking performance, enhancing male, female, Black, and White subgroup AUCs compared to unimodal, baseline multimodal, and fairness-aware models. Although CGGM reports a slightly higher AUC for the Asian subgroup, MultiFair moderately adjusts the Asian subgroup performance to reduce imbalance while strengthening performance in other racial groups, reflecting its fairness-aware modulation strategy.}

\rrevise{Table \ref{fairCLIP_dataset} presents results on the FairCLIP \cite{luo2024fairclip} dataset over six random experiments, with paired significance test. MultiFair also outperforms in terms of AUC and ES-AUC for gender (AUC of 91.25\% and ES-AUC of 89.45\%) and race subgroups (AUC of 91.51\%) and ES-AUC of 88.36\%). Additionally, Table~\ref{fairCLIP_dataset} shows that the gains in AUC and ES-AUC achieved by MultiFair are statistically significant compared to the competitive baseline models. MultiFair also improves AUCs for male, female, Asian, Black, and White subgroups compared to unimodal, baseline multimodal, and fairness-aware models. }

\rrevise{Table \ref{chexpert_unimodal_gender} summarizes the performance and fairness evaluation on the CheXpert\cite{rajpurkar2017chexpert} dataset. The results are reported over multiple random experiments, with paired statistical significance tests conducted to compare MultiFair against competitive models. MultiFair and the comparative multimodal models are trained using two modalities (Frontal and Lateral chest X-ray images) under a missing-modality scenario and a multiclass classification setup. Since the dataset comprises \myrevise{13} distinct classes, performance is reported using macro-averaged metrics (training strategy in Appendix \ref{multiclass_multimodal}), \myrevise{while additional dataset insights and training behavior analyses are presented in Appendix \ref{dataset_behavior} to better interpret the results.}   Our proposed model achieves the best overall macro AUC (80.73\%) and ES-AUC (79.82\%) among all compared methods. It consistently improves subgroup macro AUCs for both male (80.54\%) and female (80.93\%) patients compared to unimodal, baseline multimodal, and fairness-aware models.}

\rrevise{Across \cref{fairvision_result,fairCLIP_dataset,chexpert_unimodal_gender}, MultiFair does not consistently obtain the lowest DPD or DeOdds values for gender or race subgroups. \myrevise{The significan\revision{ce} test shows that in many cases, MultiFair's DPD and DeOdds results are not statistically significant compared to the other baseline models.} For example, in Table \ref{fairvision_result}, FairVision reports lower race DPD/DeOdds, and FairCLIP achieves lower gender disparity metrics. Additionally, some unimodal, baseline multimodal, and balanced multimodal models exhibit smaller DPD or DeOdds values despite having noticeably lower AUC and ES-AUC performance. A similar pattern is observed in Tables \ref{fairCLIP_dataset} and \ref{chexpert_unimodal_gender}, where certain competing methods achieve lower threshold-based parity gaps (e.g. DPD, DeOdds) while underperforming in overall ranking metrics.} \rrevise{This behavior is consistent with the design of MultiFair. The framework optimizes ranking-based fairness through AUC and ES-AUC modulation rather than directly minimizing threshold-dependent disparity metrics. Since DPD and DeOdds depend on predictions at a fixed operating threshold, improvements in ranking fairness do not necessarily result in the lowest parity gaps. Instead, MultiFair prioritizes balanced subgroup discrimination (AUCs) performance while maintaining decent disparity levels, reflecting a trade-off between ranking optimization and strict threshold-level parity.}

\textbf{Modality-Wise performance.}
When comparing across modalities, unimodal models like 3D OCT and text-only models achieve reasonable performance but exhibit clear fairness gaps. Baseline multimodal fusion increases accuracy yet fails to reduce subgroup disparities, while fairness-specific baselines like FairVision and FairCLIP reduce disparities but sacrifice prediction performance. Balanced multimodal methods partially address modality imbalance, but their improvements remain limited. In contrast, \text{\modelname} simultaneously improves subgroup balance and overall classification accuracy, reducing gaps between female and male groups and achieving more consistent outcomes across Asian, Black, and White patients.

\rrevise{Overall, MultiFair consistently achieves the highest AUC and ES-AUC across FairVision, FairCLIP, and CheXpert datasets, with statistically significant improvements over competitive baselines across six random experiments. The model enhances subgroup AUCs for both gender and race subgroups, improving performance for underrepresented groups while maintaining balanced ranking performance (e.g. AUC, ES-AUC). Although MultiFair does not always obtain the lowest DPD or DeOdds values, this aligns with its design, as it optimizes threshold-}\rrevise{independent ranking fairness rather than directly minimizing threshold-dependent parity metrics. Therefore, MultiFair achieves a strong and stable trade-off between predictive accuracy, subgroup fairness, and modality balance.}

\subsubsection{Sensitivity Analysis}
Sensitivity at 90\% Specificity($Se@Sp_{0.90}$) results are presented in Tables \ref{tab_fairvision}, \ref{tab_fairclip}  and \ref{tab:chexpert_sens90_gender}  over six random experiments (Mean ± Standard Deviation) for the FairVision, FairCLIP, and CheXpert datasets, respectively. For this analysis, we include only the most competitive models from each methodological category listed in Tables \ref{fairvision_result}, \ref{fairCLIP_dataset}, and \ref{chexpert_unimodal_gender}, where the full set of evaluated models is reported. \myrevise{In addition to subgroup sensitivities, we report the corresponding subgroup gap estimates with 95\% confidence intervals (CI) to better characterize the statistical uncertainty of disparity across the experiments.}

Table \ref{tab_fairvision} reports $Se@Sp_{0.90}$ on the FairVision dataset for the subgroups. While several competitive models achieve strong sensitivity, they show notable gender and race disparities (up to 8.4\%). \myrevise{In contrast, MultiFair attains the smallest mean subgroup gaps for both gender and race, with gap estimates of 3.10\% and 1.68\%, respectively, together with relatively narrow CI gaps.
Meanwhile, it maintains competitive subgroup sensitivities. In particular, MultiFair improves sensitivity for Females, Blacks, and Whites while only slightly reducing sensitivity for Males and Asians, yielding more balanced subgroup performance.}

\begin{table}[h]
\centering
\caption{\rrevise{Sensitivity at 90\% specificity over multiple random experiments on FairVision dataset \myrevise{ with 95\% CI Gaps.}}}
\label{tab_fairvision}
\resizebox{\columnwidth}{!}{
\color{black}
\begin{tabular}{lccccccc}
\toprule
\multirow{2}{*}{\textbf{Model}} 
& \multicolumn{3}{c}{\textbf{Gender}} 
& \multicolumn{4}{c}{\textbf{Race}} \\
\cmidrule(lr){2-4}
\cmidrule(lr){5-8}
& \textbf{Male} & \textbf{Female} & \textbf{\myrevise{95\% CI Gap}} 
& \textbf{Asian} & \textbf{Black} & \textbf{White} & \textbf{\myrevise{95\% CI Gap}} \\
\midrule

2D (VGG16)        & 52.49±3.26 & 48.58±3.30 & \myrevise{[0.51, 7.30]} & 52.03±6.47 & 48.28±7.93 & 50.65±1.81 & \myrevise{[1.77, 11.87]} \\
3D (ResNet18)     & 64.82±2.01 & 58.30±1.91 & \myrevise{[5.81, 9.31]}
 & 59.83±5.31 & 63.81±3.09 & 60.72±1.69 & \myrevise{[0.03, 8.35]} \\

Baseline (ResNet18)        & 62.11±2.52 & 56.60±1.90 & \myrevise{[4.65, 6.37]} & 61.19±5.24 & 62.46±3.91 & 57.98±2.55 & \myrevise{[4.28, 9.83]} \\
FairCLIP                   & 69.92±2.68 & 61.48±1.65 & \myrevise{[6.63, 10.26]} & 63.39±2.50 & 66.72±4.00 & 65.09±2.32 & \myrevise{[0.73, 9.55]} \\
MultiViT                   & 67.54±2.17 & 59.14±3.15 & \myrevise{[6.14, 10.67]} & 62.37±4.51 & 64.78±3.89 & 62.48±2.47 & \myrevise{[0.28, 7.34]
} \\
CGGM                       & 69.22±2.93 & 62.47±2.25 & \myrevise{[2.17, 11.33]} & 68.64±3.39 & 67.09±3.73 & 64.74±2.35 & \myrevise{[1.14, 9.01]} \\
MultiFair                  & 68.43±4.29 & 65.21±3.58 & \myrevise{[1.79, 5.76]} & 66.76±4.66 & 67.66±1.32 & 65.98±1.89 & \myrevise{[0.77, 4.59]} \\
\bottomrule
\end{tabular}
}
\end{table}

\begin{table}[h]
\centering
\caption{\rrevise{Sensitivity at 90\% specificity over multiple random experiments on FairCLIP dataset \myrevise{ with 95\% CI Gaps.}}}
\label{tab_fairclip}
\resizebox{\columnwidth}{!}{
\color{black}
\begin{tabular}{lccccccc}
\toprule
\multirow{2}{*}{\textbf{Model}} 
& \multicolumn{3}{c}{\textbf{Gender}} 
& \multicolumn{4}{c}{\textbf{Race}} \\
\cmidrule(lr){2-4}
\cmidrule(lr){5-8}
& \textbf{Male} & \textbf{Female} & \textbf{\myrevise{95\% CI Gap}}
& \textbf{Asian} & \textbf{Black} & \textbf{White} & \textbf{\myrevise{95\% CI Gap}} \\
\midrule

2D (VGG16)              & 58.65$\pm$1.99 & 53.52$\pm$1.01 & \myrevise{[1.86, 8.38]}
 & 60.25$\pm$6.55 & 56.53$\pm$3.70 & 55.16$\pm$0.78 & \myrevise{[2.42, 12.34]
} \\
Text (Google BERT)      & 65.44$\pm$3.10            & 60.47$\pm$2.25            & \myrevise{[5.15, 9.66]}  & 66.21$\pm$1.20            & 60.33$\pm$2.10            & 59.11$\pm$2.1            & \myrevise{[5.21, 13.82]}  \\
Baseline (EfficientNet) & 66.94$\pm$2.87 & 63.04$\pm$1.77 & \myrevise{[1.16, 6.64]} & 69.87$\pm$5.47 & 67.04$\pm$2.76 & 63.66$\pm$2.07 & \myrevise{[2.71, 11.58]} \\
FairVision              & 67.82$\pm$2.65 & 63.33$\pm$3.48 & \myrevise{[2.45, 6.53]} & 59.24$\pm$5.62 & 67.35$\pm$5.31 & 61.90$\pm$3.67 & \myrevise{[4.80, 13.29]} \\
BLIP-2                  & 58.65$\pm$2.62 & 54.09$\pm$2.01 &  \myrevise{[2.23, 6.89]} & 60.51$\pm$2.59 & 56.94$\pm$3.79 & 55.45$\pm$2.84 & \myrevise{[3.77, 10.13]}\\
OGM                     & 70.48$\pm$0.88 & 65.03$\pm$1.63 & \myrevise{[4.42, 6.49]} & 72.91$\pm$3.42 & 68.06$\pm$2.52 & 66.74$\pm$1.29 & \myrevise{[2.36, 10.60]} \\
MultiFair               & 72.98$\pm$1.84 & 69.88$\pm$1.53 & \myrevise{[1.09, 5.44]} & 75.58$\pm$4.72 & 71.38$\pm$3.65 & 69.64$\pm$1.42 & \myrevise{[3.74, 8.74]} \\
\bottomrule
\end{tabular}
}
\end{table}

Table \ref{tab_fairclip} illustrates $Se@Sp_{0.90}$ on the FairClIP dataset across demographic subgroups. MultiFair shows the minimum average sensitivity gap (3.10\%) for the gender subgroups. Although the mean gap (5.94\%) in the race subgroups is not the lowest among the compared methods, MultiFair consistently increases sensitivity across all racial groups (Asian, Black, and White) relative to the other models. \myrevise{The reported 95\% CI further indicates that the subgroup disparities remain controlled while sensitivity is improved.
}

Table \ref{tab:chexpert_sens90_gender} reports macro-averaged $Se@Sp_{0.90}$ over 13 pathologies on the CheXpert dataset. While several methods achieve relatively small gender gaps, their overall sensitivity remains limited. In contrast, MultiFair attains the highest sensitivity for both male (46.93\%) and female (46.60\%) subgroups while maintaining a comparatively low gender gap (0.33\%) \myrevise{with a tight CI.
These results demonstrate that MultiFair improves diagnostic sensitivity without introducing substantial subgroup disparity. 
}
\vspace{-3mm}
\begin{table}[h]
\centering
\caption{\rrevise{Sensitivity at 90\% specificity macro-averaged over 13 pathologies on CheXpert dataset \myrevise{ with 95\% CI Gaps.}}}
\label{tab:chexpert_sens90_gender}
\resizebox{0.6\columnwidth}{!}{
\color{black}
\begin{tabular}{lccc}
\toprule
\multirow{2}{*}{\textbf{Model}} & \multicolumn{3}{c}{\textbf{Gender}} \\
\cmidrule(lr){2-4}
& \textbf{Male} & \textbf{Female} & \textbf{\myrevise{95\% CI Gap}} \\
\midrule
Unimodal                 & 45.68$\pm$0.29 & 44.62$\pm$0.89 & \myrevise{[0.26, 1.86]} \\
Multimodal               & 45.35$\pm$1.17 & 44.34$\pm$1.25 & \myrevise{[0.20, 1.99]} \\
FairCLIP                 & 36.29$\pm$1.55 & 35.97$\pm$2.16 & \myrevise{[0.01, 1.24]} \\
MoE                      & 38.87$\pm$1.87 & 38.02$\pm$1.16 & \myrevise{[0.25, 2.09]} \\
OPM                      & 40.58$\pm$0.52 & 40.67$\pm$0.65 & \myrevise{[0.02, 0.60]} \\
MultiFair   & 46.93$\pm$1.54 & 46.60$\pm$1.27 & \myrevise{[0.05, 0.70]} \\
\bottomrule
\end{tabular}}
\end{table}

\vspace{-2mm}
Overall, sensitivity analysis across three datasets signifies that MultiFair enhances subgroup equity without sacrificing overall diagnostic performance. \myrevise{CIs for subgroup gaps further strengthen this conclusion by showing that the observed fairness improvements are not only competitive in magnitude but also stable across repeated random experiments.}

\subsubsection{Balanced Subgroup Performance}
\rrevise{To evaluate the robustness of MultiFair under balanced subgroup distributions, we conducted additional experiments using resampling strategies. The original datasets (FairVision and FairCLIP) exhibit demographic imbalance, with a gender distribution of 57.0\% female and 43.0\% male, and a racial distribution of 8.5\% Asian, 14.9\% Black, and 76.6\% White based on self-reported information. To mitigate this imbalance, we construct balanced training, validation, and test sets using two strategies. First, \textit{upsampling} is applied to minority subgroups by sampling with replacement} \rrevise{until all subgroup counts match the largest group. Second, \textit{downsampling} is performed by randomly sampling from the majority groups (with replacement) to match the size of the smallest subgroup. These procedures are applied separately for} \rrevise{gender and race attributes. The same demographic distribution and resampling protocol are used for both the FairVision and FairCLIP datasets.}

\begin{table*}[t]
\centering
\caption{\rrevise{Performance of MultiFair under different resampling strategies on FairVision and FairCLIP datasets. }}
\label{sampling}
\resizebox{0.8\textwidth}{!}{
\color{black}
\begin{tabular}{llccccccccc}
\toprule
 &  & \multicolumn{4}{c}{\textbf{Gender}} & \multicolumn{5}{c}{\textbf{Race}} \\
\cmidrule(lr){3-6} \cmidrule(lr){7-11}
\textbf{Dataset} & \textbf{Sampling} & \textbf{AUC} & \textbf{ES-AUC} & \textbf{Male} & \textbf{Female} & \textbf{AUC} & \textbf{ES-AUC} & \textbf{Asian} & \textbf{Black} & \textbf{White} \\
\midrule

\multirow{3}{*}{FairVision}
& MultiFair 
  & 86.07±0.57 & 83.03±0.25 & 88.05±0.84 & 84.39±0.35
  & 86.31±0.36 & 82.02±0.48 & 89.04±1.13 & 83.15±0.72 & 85.97±0.45 \\
& $\text{MultiFair}_{\text{up}}$ 
& 86.20±0.48 & 83.28±0.37 & 86.94±2.04 & 84.06±0.47
& 86.58±0.51 & 81.23±0.96 & 88.91±1.43 & 83.41±0.65 & 85.48±0.59 \\
& $\text{MultiFair}_{\text{down}}$
& 85.90±0.61 & 82.58±0.65 & 86.87±1.92 & 83.47±0.42
& 86.65±0.57 & 81.18±1.31 & 89.12±1.34 & 84.03±0.51 & 84.99±0.86 \\

\midrule

\multirow{3}{*}{FairCLIP}
& MultiFair
    & 91.25±0.44 & 89.45±0.63 & 92.37±0.42 & 90.32±0.55 
    & 91.51±0.41 & 88.36±0.85 & 94.36±0.67 & 91.08±0.83 & 91.33±0.39 \\
& $\text{MultiFair}_{\text{up}}$
& 91.30±0.59 & 89.58±0.36 & 92.27±1.05 & 90.32±0.18
& 92.34±0.57 & 86.01±0.97 & 95.60±0.35 & 90.00±0.50 & 89.31±0.0049 \\
& $\text{MultiFair}_{\text{down}}$
& 90.41±0.38 & 88.42±0.76 & 91.50±0.94 & 89.24±0.35
& 90.80±0.50 & 85.59±1.13 & 93.56±0.69 & 87.61±0.90 & 88.51±0.61 \\

\bottomrule
\end{tabular}}
\end{table*}

\rrevise{Table~\ref{sampling} summarizes the results for three settings: \textit{MultiFair}, which trains on the original imbalanced dataset; $\text{MultiFair}_{\text{up}}$, which trains with subgroup upsampling; and $\text{MultiFair}_{\text{down}}$, which denotes training with} \rrevise{subgroup downsampling. Results are reported as Mean $\pm$ Standard Deviation for 3 independent random experiments. Overall, the results show that \textit{MultiFair} maintains comparable performance under balanced subgroup distributions. Both upsampling and downsampling yield comparatively similar (results on downsampling show a slightly lower performance) AUC and ES-AUC values across gender and race groups, indicating that the proposed framework remains stable when subgroup imbalance is mitigated. These findings suggest that the effectiveness of \textit{MultiFair} is not dependent on the original demographic distribution and that the model can perform consistently even when subgroup proportions are artificially balanced.}

\color{black}
\subsection{Computational Overhead}
\rrevise{Table \ref{tab:computational_overhead} summarizes the computational overhead of different multimodal model categories for training and inference on FairVision\cite{luo2024fairvision}. All results are Mean ± Standard Deviation over 6 random experiments with a batch size of 32. The dataset comprises 10,000 samples,
with each sample containing 2D SLO fundus,3D OCT B-scan volumes, and associated demographic subgroup labels. The dataset is divided into 6,000 training samples, 1,000 validation samples, and 3,000 test samples, following a 60\%, 10\%, and 30\% split. 
The model was trained on the NVIDIA A100 with 80GB of RAM.}

The epoch time and GPU memory metrics capture 
\rrevise{the additional training cost associated with model-specific operations, including modality balancing and fairness-aware gradient modulation. The proposed MultiFair \revision{framework} experiences a slight increase in training-time overhead due to dual-level modulation calculations, while remaining substantially more efficient than balanced multimodal baselines that employ heavier gradient-level operations. Importantly, dual-level modulation in MultiFair is applied only during training; therefore, no additional computation is introduced at inference, and the inference time remains comparable to standard multimodal and fairness-aware baselines.}

\begin{table}[t]
\centering
\caption{\rrevise{Computational overhead comparison across different multimodal model categories on FairVision dataset.}}
\label{tab:computational_overhead}
\footnotesize
\begin{adjustbox}{max width=0.47\textwidth}
\color{black}
\begin{tabular}{llcccc}
\toprule
\multirow{2}{*}{\textbf{Category}} &
\multirow{2}{*}{\textbf{Model}} &
\multicolumn{4}{c}{\textbf{Overhead Metrics}} \\
\cmidrule(lr){3-6}
& &
\textbf{Epoch} &
\textbf{GPU} &
\textbf{GPU} &
\textbf{Infer.} \\
& &
\textbf{Time (s)} &
\textbf{Alloc. (GB)} &
\textbf{Reserve (GB)} &
\textbf{Time (s)} \\
\midrule

\multirow{4}{*}{Baseline}
& ViT           & 339.69 ± 85.67 & 11.18 ± 0.71 & 15.03 ± 1.15 & 95.97 ± 2.79 \\
& EfficientNet  & 219.21 ± 23.48 & 6.67 ± 0.02  & 8.28 ± 0.01  & 84.16 ± 6.17 \\
& ResNet        & 189.25 ± 3.95  & 3.64 ± 0.08  & 4.74 ± 0.01  & 89.55 ± 1.03 \\
& VGG           & 216.43 ± 14.51 & 10.56 ± 0.96 & 13.05 ± 0.71 & 89.81 ± 2.96 \\
\midrule

\multirow{2}{*}{Fairness}
& FairVision    & 266.49 ± 4.37  & 11.74 ± 0.43 & 15.47 ± 0.94 & 93.11 ± 0.86 \\
& FairCLIP      & 273.32 ± 49.41 & 7.60 ± 0.05  & 13.88 ± 0.00 & 85.65 ± 13.87 \\
\midrule

\multirow{2}{*}{Existing}
& CrossViT      & 591.97 ± 10.07 & 43.98 ± 0.36 & 45.58 ± 0.40 & 133.63 ± 8.86 \\
& MultiViT      & 590.12 ± 6.50  & 43.77 ± 0.29 & 44.98 ± 0.40 & 138.31 ± 1.96 \\
\midrule

\multirow{3}{*}{Balanced}
& OPM           & 721.81 ± 29.65 & 77.53 ± 0.32 & 78.28 ± 0.52 & 168.47 ± 16.33 \\
& OGM           & 558.19 ± 17.51 & 77.55 ± 0.01 & 78.39 ± 0.00 & 135.68 ± 14.27 \\
& CGGM          & 79.63 ± 2.17   & 28.96 ± 0.38 & 30.15 ± 0.33 & 74.93 ± 3.37 \\
\midrule

\multirow{1}{*}{Proposed}
& MultiFair     & 272.53 ± 4.24  & 30.84 ± 0.94 & 32.16 ± 0.92 & 74.45 ± 3.19 \\
\bottomrule
\end{tabular}
\end{adjustbox}
\end{table}

\begin{figure}[!t]
  \centering
  \subfloat[]{
    \includegraphics[width=0.45\linewidth]{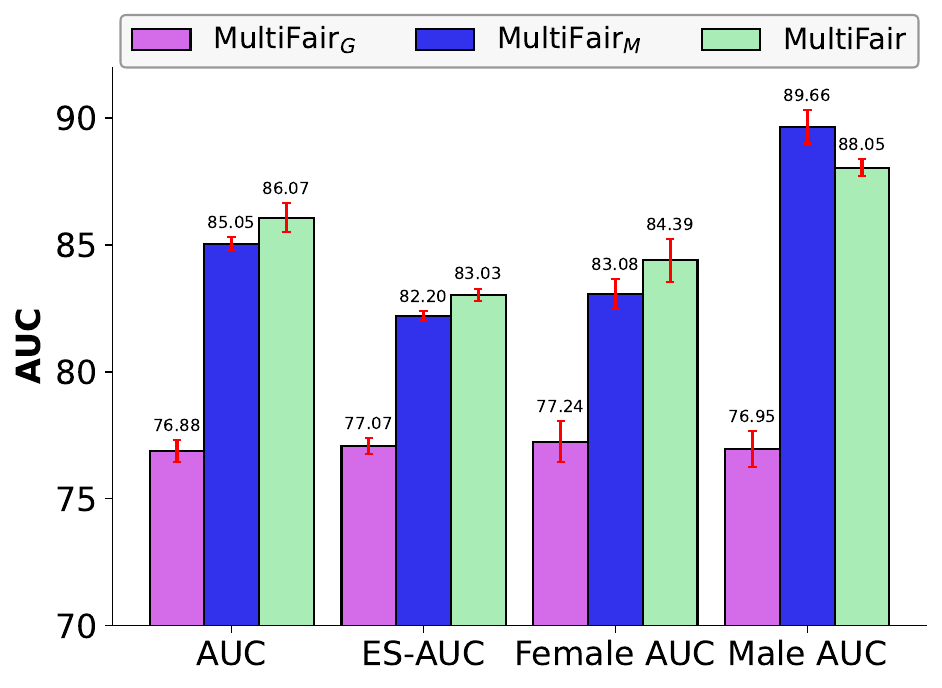}\label{gender}}
  \hfill
  \subfloat[]{
    \includegraphics[width=0.52\linewidth]{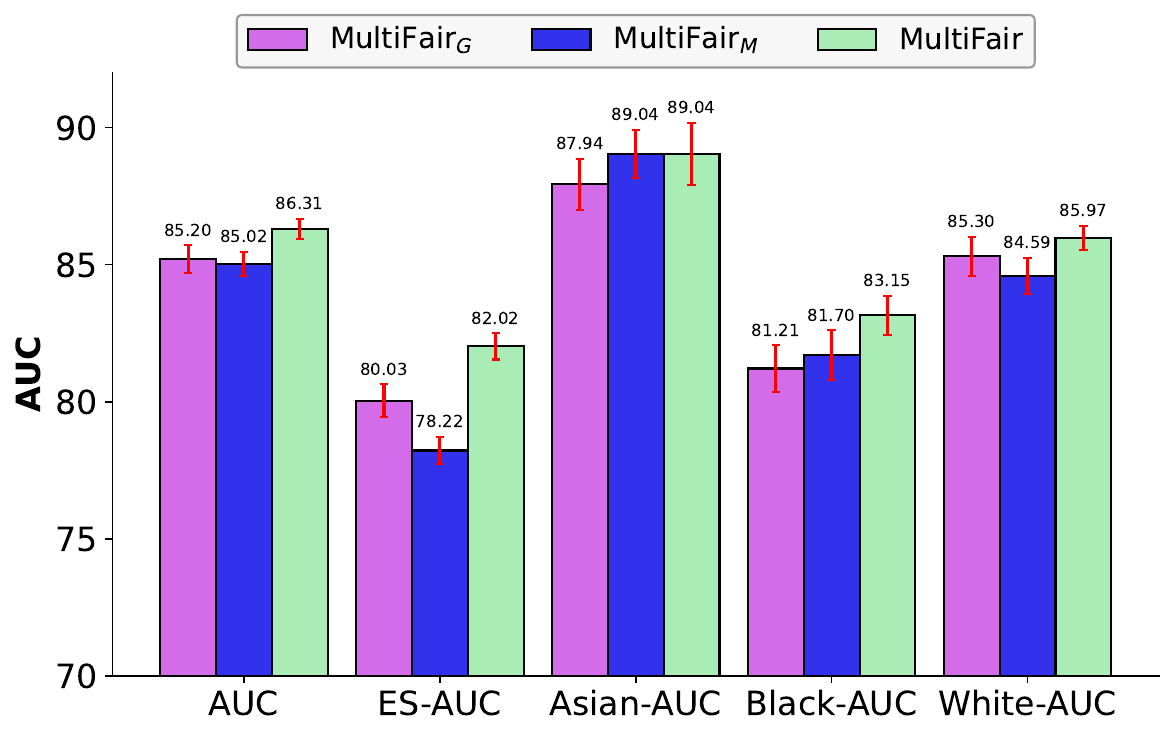}\label{race}}
    \vspace{-3mm}
    \caption{\textbf{Ablation study results}. \rrevise{We represent the performance of fairness only modulation with \text{\modelname}$_G$, modality only modulation with \text{\modelname}$_M$, and dual modulation with \text{\modelname} for the MultiFair framework. The figure (a) represents the performance for the gender groups, and (b) shows the corresponding performance for various racial subgroups.}}
  \label{fig_ablation}
  \vspace{-5mm}
\end{figure}

\color{black}
\subsection{Ablation Study}
\label{ablation}
We evaluate \text{MultiFair} under fairness-only modulation \rrevise{(excluding the modality modulation)}, modality-only modulation \rrevise{(excluding fairness-aware modulation)}, and combined dual-level modulation to investigate their relative contributions to predictive performance and fairness. \rrevise{Fig. \ref{fig_ablation}} summarizes the results for gender and race subgroups on the FairVision \cite{luo2024fairvision} dataset.\rrevise{We use the same parameter settings as described in the section \ref{experimental_settings} for the ablation study. Results are reported over six independent random experiments, where the bars represent the mean AUC values and the vertical error bars indicate the standard deviation across the six runs.} 

\rrevise{For gender (Fig.~\ref{gender}), fairness-only modulation (MultiFair$_G$) yields relatively balanced Male and Female AUCs (approximately 77\%), but with reduced overall AUC and ES-AUC. In contrast, modality-only modulation (MultiFair$_M$) substantially improves overall AUC and ES-AUC. However, it introduces a noticeable disparity between Male and Female AUCs (around 6--7\%). When both fairness and modality modulation are jointly applied (MultiFair), Female AUC increases compared to modality-only modulation, while Male AUC is moderately adjusted. MultiFair achieves \revision{the} best overall AUC and ES-AUC, demonstrating the effectiveness of both modulation strategies in balancing predictive performance and fairness. }

\rrevise{For the racial subgroups (Fig.~\ref{race}), fairness-only and modality-only modulation achieve comparable overall AUC but show different subgroup AUCs. Modality-only modulation produces a higher Asian AUC, whereas fairness-only modulation yields more uniform subgroup performance with lower ES-AUC. By combining both modulation mechanisms, MultiFair moderates the disproportionately high Asian AUC while improving Black and White subgroup AUCs. This} \rrevise{leads to improved subgroup balance and the highest ES-AUC with competitive overall AUC, demonstrating that the joint modulation strategy is necessary for stable and fairness-aware performance across racial groups. We also include additional ablation studies on the FairCLIP dataset in Appendix \ref{additional_ablation}.}

\subsection{Parameter Sensitivity}
\begin{figure*}[h]
  \centering
  \subfloat[]{
    \includegraphics[width=0.235\linewidth]{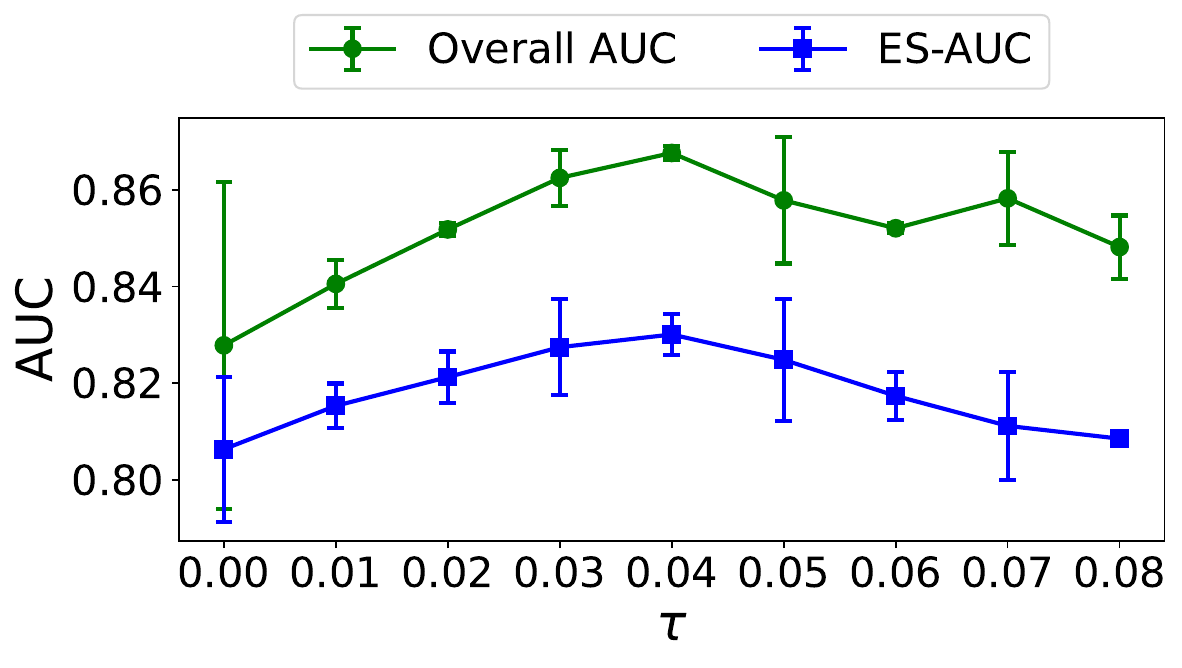}\label{tau}}
  \hfill
  \subfloat[]{
    \includegraphics[width=0.235\linewidth]{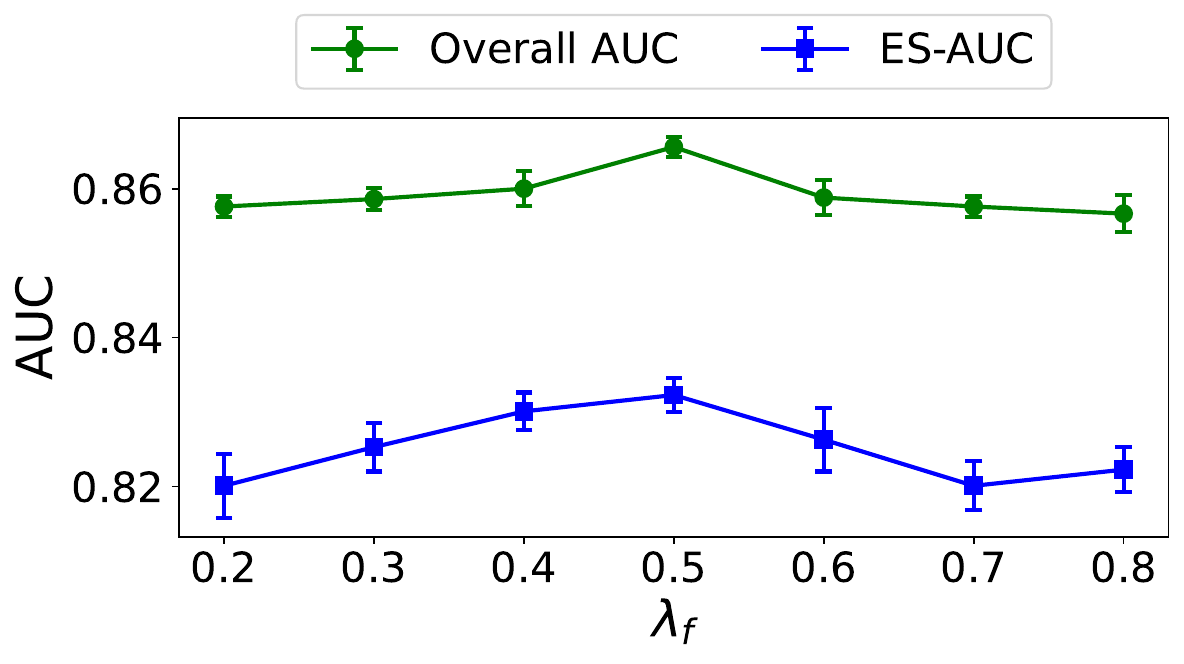}\label{lamda}}
  \hfill
  \subfloat[]{
    \includegraphics[width=0.235\linewidth]{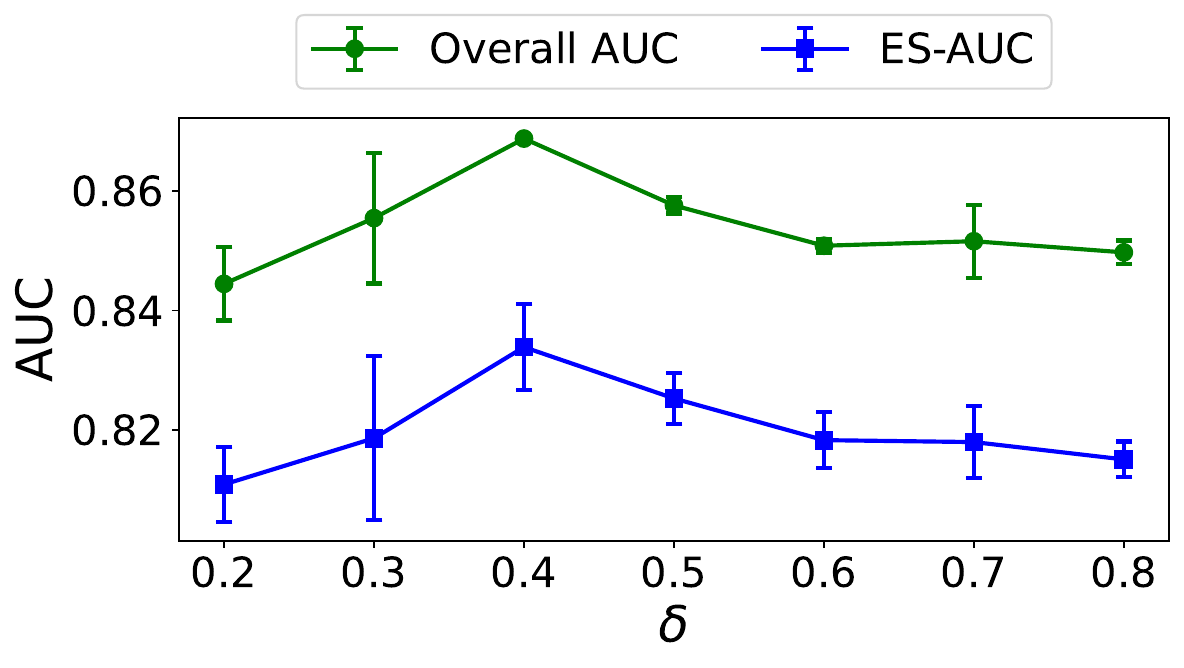}\label{delta}}
  \hfill
  \subfloat[]{
    \includegraphics[width=0.235\linewidth]{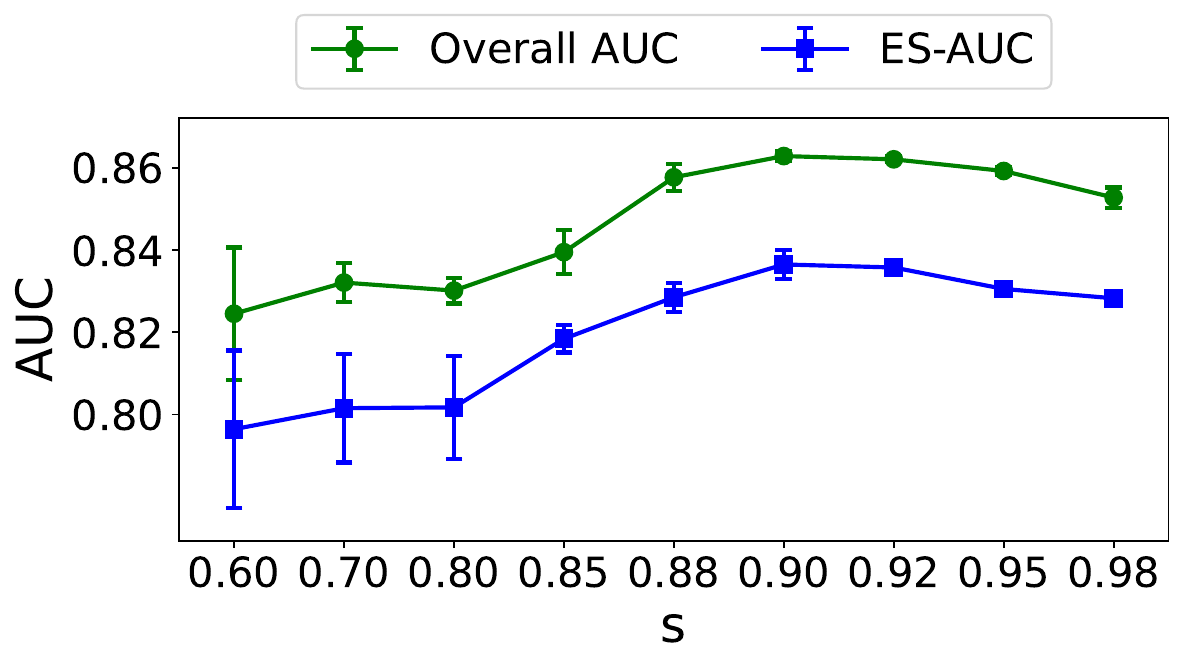}\label{ema_sweep}}
  
    \vspace{-1mm}
  \caption{\textbf{The impact of different fairness parameters}. (a) shows the variation in AUCs across different fairness thresholds $(\tau)$. (b) shows the effect of the fairness penalty $\lambda_f$. (c) illustrates how the performance changes with the various values of fairness modulation strength $(\delta)$. \revision{(d) shows the impact of the EMA smoothing factor s.} \rrevise{The results are reported over three independent experiments on FairVision dataset. The horizontal lines indicate the mean AUC values, while the vertical bars represent the corresponding standard deviations.} } 
  \label{parameters}
  \vspace{-5mm}
\end{figure*}
\rrevise{MultiFair involves two key categories of hyperparameters: modality balancing and fairness optimization. The guidelines for selecting the optimal values of both sets are provided in Appendix~\ref{hyperparameter_selection}. Since our modality balancing strategy follows the procedure of the CGGM model~\cite{guo2024classifier}, we adopt their recommended guidelines for tuning the modality balancing hyperparameters ($\rho$ and $\lambda_{gm}$).}

\color{black}
Fig.~\ref{parameters} illustrates the impact of the fairness hyperparameters on overall AUC and ES-AUC. As shown in Fig.~\ref{tau}, performance initially improves with increasing fairness threshold ($\tau$) and reaches a stable optimum around $\tau = 0.04$, beyond which both AUC and ES-AUC gradually decline. In Fig.~\ref{lamda}, moderate values of the fairness penalty ($\lambda_f = 0.5$) provide the best trade-off between performance and fairness, while excessively large penalties slightly degrade the performance. Similarly, Fig.~\ref{delta} shows that the fairness modulation strength $\delta = 0.4$ yields the highest AUC and ES-AUC, whereas both smaller and larger values reduce stability. \revision{Fig. \ref{ema_sweep} evaluates the EMA smoothing factor $s$, showing that moderate values provide the best balance between stability and responsiveness in subgroup surrogate-AUC estimates. Performance improves as $s$ increases, remains stable between 0.88 and 0.92, and reaches its best value at s=0.90, whereas excessively large values slightly reduce performance due to over-smoothing.}

\rrevise{Based on these observations, we select intermediate values of $s$, $\tau$, $\lambda_f$, and $\delta$ that maximize ES-AUC while maintaining stable overall AUC, ensuring an effective balance between fairness regularization and predictive performance.}

\begin{figure}[!t]
  \centering
  \subfloat[]{
    \includegraphics[width=0.48\linewidth]{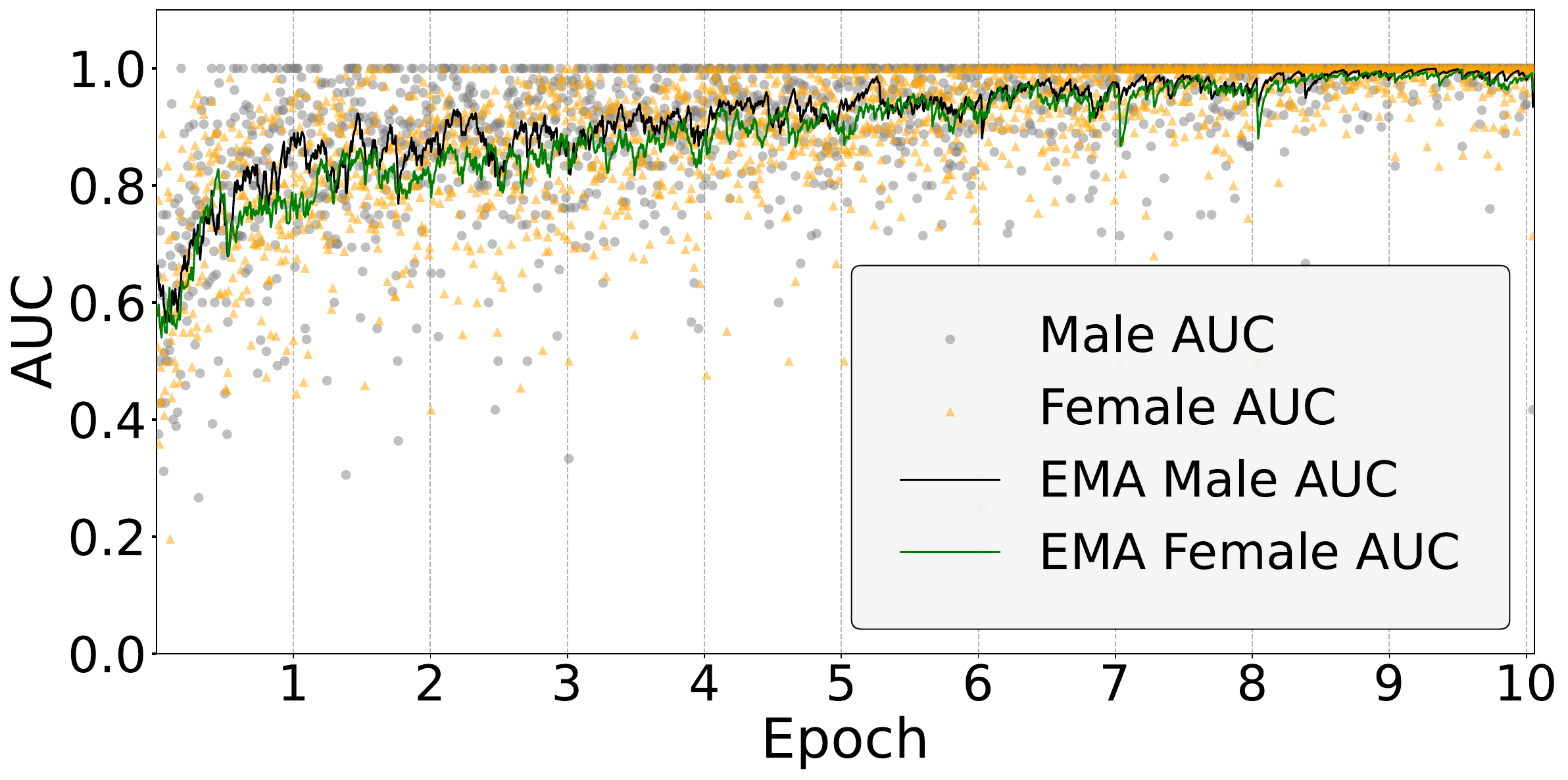}\label{batch_32}}
  \hfill
  \subfloat[]{
    \includegraphics[width=0.48\linewidth]{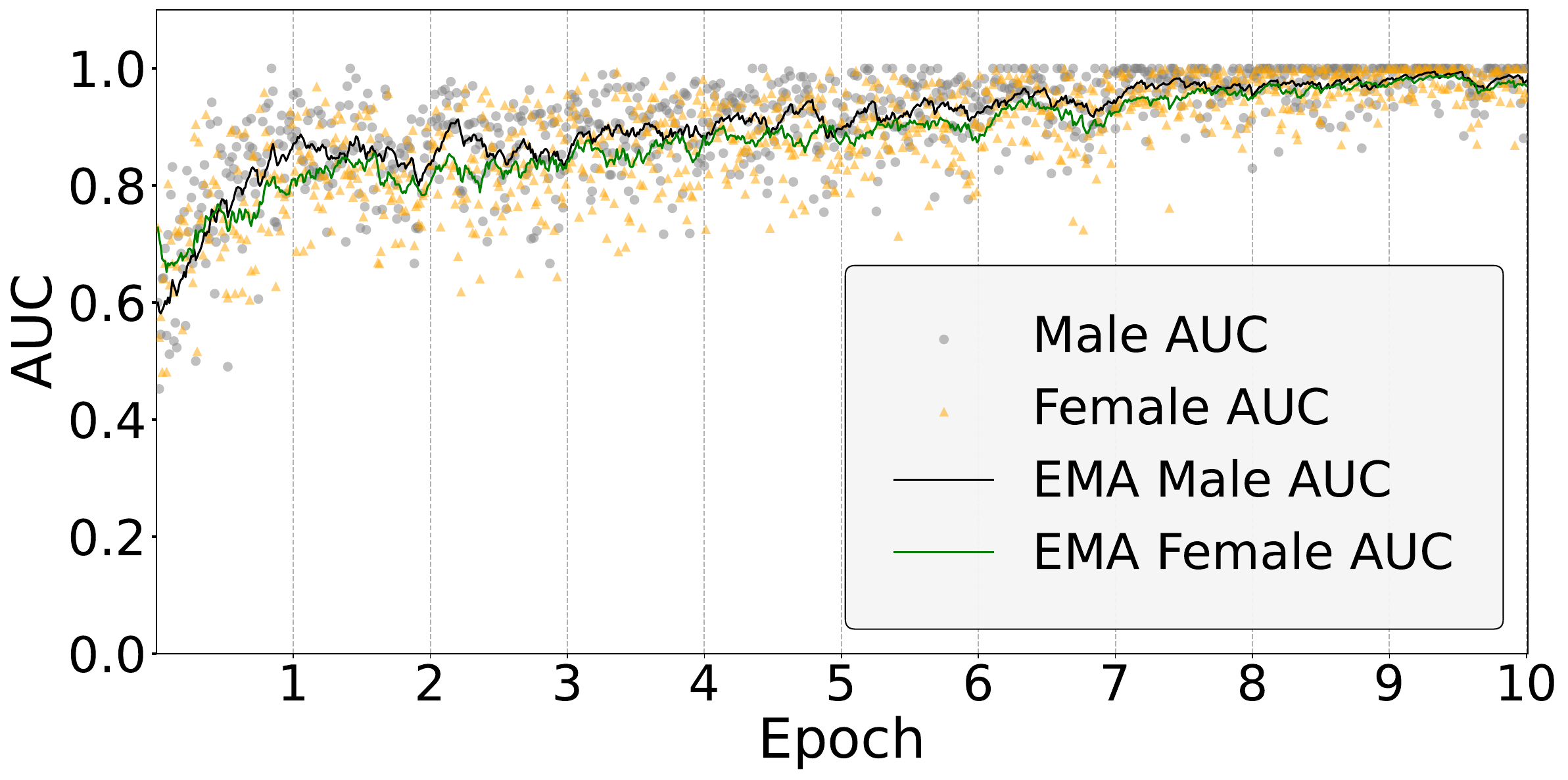}\label{batch_64}}
    \vspace{-1mm}
  \caption{\textbf{Surrogate AUC stability with EMA smoothing.} Convergence of batch-level surrogate AUC for male and female subgroups, comparing raw estimates and EMA-smoothed estimates (s = 0.9). Results are shown on FairVision Dataset for (a) batch size 32 and (b) batch size 64.}
  \label{fig_batch_ema}
  \vspace{-5mm}
\end{figure}

\rrevise{\subsection{Surrogate AUC stability with EMA smoothing.} \label{AUC_stability}
During the training, we used exponential moving average (EMA) of surrogate AUC instead of raw surrogate AUC according to Eq. \ref{ema}. Figs. \ref{fig_batch_ema} show that batch-level surrogate AUC estimates exhibit
substantial stochastic variability under subgroup splits, particularly at batch size 32 (Fig. \ref{batch_32}) due to limited subgroup samples within mini-batches. Increasing the batch size to 64 (Fig. \ref{batch_64}) reduces this variance but does not fully eliminate early-training noise. On the contrary, EMA smoothing $(s=0.9)$ consistently stabilizes subgroup AUC trajectories across both batch sizes,} \rrevise{providing reliable estimates for fairness-aware gradient modulation.}

\begin{figure}
  \centering
    \includegraphics[width=0.45\textwidth]{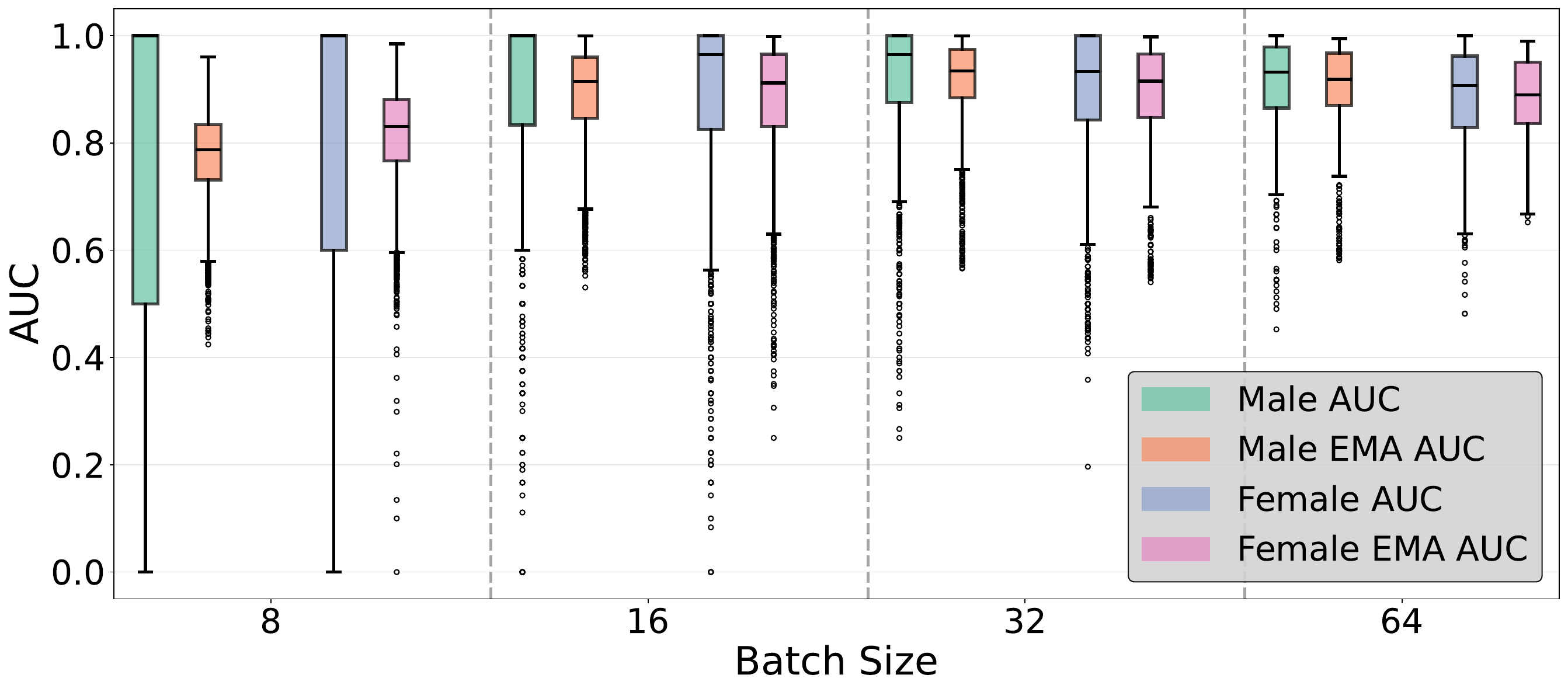}
  \caption{\textbf{Distribution of batch-level surrogate AUC across batch sizes}. Boxplots show that batch-level surrogate AUC for male and female subgroups shows high variance at small batch sizes (8 and 16) in raw estimates, while increasing batch sizes (32 and 64) show relatively low variance. Applying EMA smoothing (s = 0.9) significantly reduces variability and yield\revision{s} stable subgroup AUC estimates.}
  \label{boxplot_batch_size}
  \vspace{-4mm}
\end{figure}

\rrevise{Fig. \ref{boxplot_batch_size} illustrates the batch-level surrogate AUC variance on different batch sizes under subgroup splits. The model was trained on the FairVision \cite{luo2024fairvision} dataset. Raw} \rrevise{surrogate AUC shows large dispersion at small batch sizes due to sparse subgroup samples. Although increasing batch size reduces variance, noticeable noise remains even at batch size 32. EMA smoothing consistently stabilizes subgroup AUC estimates, yielding tight distributions and preventing fairness modulation from reacting to transient batch-level noise.}

\section{Discussion}
\color{black}
MultiFair balances the modalities and ensures fairness across groups for multimodal medical classification tasks. It uses both gradient magnitude and direction to balance modalities, and subgroup fairness disparities to adjust the gradients accordingly. \myrevise{Additionally, it optimizes ranking-based fairness such as group AUCs and ES-AUC, but it \revision{does} not necessarily correspond to optimal threshold-based parity metrics such as DPD or DeOdds.} \color{black} The proposed framework does not assume that modality availability differs across demographic groups in the evaluated datasets. Instead, it focuses on mitigating optimization imbalance during training, where different modalities may contribute gradients of unequal magnitudes, potentially causing modality dominance. The results of the Tables \ref{fairvision_result}, \ref{fairCLIP_dataset}, \rrevise{and \ref{chexpert_unimodal_gender}} indicate the effectiveness of our model on the FairVision, FairCLIP \rrevise{and CheXpert} datasets. \rrevise{The model is evaluated across multiple modalities, including scenarios with missing modalities, for both binary and multiclass classification tasks.}

\myrevise{
However, the current evaluation is limited to two glaucoma datasets with gender and race-based fairness analyses and one chest X-ray dataset with gender-only fairness analysis. Thus, the findings indicate the effectiveness of MultiFair in the studied settings, while broader validation is needed to further ensure its generalizability.} The proposed model optimizes fairness using a surrogate AUC objective. Therefore, improvements in AUC and ES-AUC do not always produce the lowest DPD or DeOdds values, since these metrics depend on a decision threshold. \myrevise{While MultiFair demonstrates empirical stability of the fairness activation rule $\Delta AUC \ge \tau$ (Figure \ref{activation_rule}), the rule remains heuristic and its robustness may vary across settings. Future work will explore more robust alternatives, such as hysteresis-based and adaptive triggering mechanisms.} In the multiclass setting with partial modality availability, the one-vs-rest surrogate AUC is computed only on modality-present samples, which can lead to sparse pairwise comparisons and introduce estimation noise in fairness optimization. Although training stability is supported using a relatively large batch size (e.g., 128) and EMA smoothing, performance may still be sensitive to batch composition. Moreover, in CheXpert, the fairness signal is influenced mainly by the paired frontal-lateral subset, and rare pathologies may have limited subgroup support and higher fallback frequency, making some pathology-level fairness estimates less stable.  In the current formulation, MultiFair is trained separately for each sensitive attribute, resulting in distinct models for different fairness objectives. Additionally, the current framework assumes paired modalities within each patient sample and does not explicitly address unpaired or asynchronous modalities across patients.

\rrevise{Future research could incorporate threshold-aware fairness regularization to jointly optimize ranking-based and parity-based fairness criteria. Developing more robust estimation strategies for multiclass fairness under missing modalities would further improve stability. \myrevise{We plan to explore adaptive smoothing to dynamically adjust the EMA factor for improved stability and responsiveness,} \myrevise{and extend the model to support joint multi-attribute fairness optimization, enabling a unified model that balances fairness across multiple demographic attributes.} Moreover, extending MultiFair to handle unpaired multimodal learning under fairness constraints represents an important direction for enhancing its applicability in realistic clinical deployment settings. We also note that our current evaluation is limited to the studied datasets, and future work will explore broader external validation across different institutions, devices, and acquisition settings.}
\color{black}
\section{Conclusion}
We present \text{MultiFair}, a dual-level gradient modulation framework for multimodal medical classification that simultaneously addresses modality imbalance and demographic unfairness. By jointly modulating gradients at the modality and subgroup levels and integrating task accuracy, gradient alignment, and fairness gap minimization into a unified loss, MultiFair achieves balanced convergence and equitable performance. Experiments on \rrevise{three} real-world datasets show that MultiFair generally outperforms state-of-the-art unimodal, fairness-aware, and balanced multimodal models in both AUC and ES-AUC, while maintaining competitive fairness metrics. \myrevise{However, these gains are not directly related to threshold-based metrics such as DPD and DeOdds.} \myrevise{Our theoretical analysis provides insight into the stability and fairness-aware behavior of the method. Beyond medical imaging, MultiFair shows potential as a fairness-aware multimodal learning framework, although its applicability to other health and safety-critical settings \revision{is yet} to be validated.}

\appendix

\subsection{Additional Algorithm}
\label{appendix_algo}
\rrevise{We used the differentiable surrogate AUC instead of the traditional AUC for the fairness modulation as described in section \ref{fairness_balancing}. And the surrogate AUC is calculated according to the Algorithm \ref{alg:surrogate_auc}}.
\begin{algorithm}[h]
\color{black}
\caption{Pairwise Surrogate AUC}
\label{alg:surrogate_auc}
\KwIn{Mini-batch predictions $\{s_{j,i}\}$, labels $\{y_j\}$, demographic group $g$, modality $i$, temperature $T$ (set to 1)}
\KwOut{$\mathrm{AUC}^{\text{batch, (t)}}_{g,i}$}

$\mathcal{P}_{g,i}^{(t)} \leftarrow \{j \mid y_j = 1 \}$ \;
$\mathcal{N}_{g,i}^{(t)} \leftarrow \{j \mid y_j = 0 \}$ \;

\eIf{$|\mathcal{P}_{g,i}^{(t)}| = 0$ \textbf{or} $|\mathcal{N}_{g,i}^{(t)}| = 0$}{
    $\mathrm{AUC}^{\text{batch, (t)}}_{g,i} \leftarrow 0.5$ 
}{
    $\mathcal{L} \leftarrow 0$ \;
    \For{$p \in \mathcal{P}_{g,i}^{(t)}$}{
        \For{$n \in \mathcal{N}_{g,i}^{(t)}$}{
            $\mathcal{L} \leftarrow \mathcal{L}
            + \log\!\left(1 + e^{-\frac{s_{p,i} - s_{n,i}}{T}}\right)$ \;
        }
    }
    $\mathcal{L} \leftarrow
    \mathcal{L} \, / \,
    \big(|\mathcal{P}_{g,i}^{(t)}|
    |\mathcal{N}_{g,i}^{(t)}|\big)$ \;
    $\mathrm{AUC}^{\text{batch, (t)}}_{g,i} \leftarrow 1 - \mathcal{L}$ \;
}
\Return $\mathrm{AUC}^{\text{batch, (t)}}_{g,i}$
\end{algorithm}

\subsection{Guideline for training MultiFair}
\subsubsection{Hyperparameter Selection}
\label{hyperparameter_selection}
\rrevise{ The proposed MultiFair framework introduces a dual modulation strategy that addresses modality imbalance and fairness across demographic subgroups. For modality balancing, the key hyperparameters are the modality magnitude scaling factor ($\rho$) and the direction modulation coefficient ($\lambda_{gm}$). To ensure stable and effective optimization, both hyperparameters should be selected within moderate ranges. Empirically, $\rho$ performs well within $[1.0$--$1.3]$, with $\rho = 1.2$ serving as a practical default. Similarly, $\lambda_{gm}$ is generally effective within $[0.10$--$0.20]$, with $\lambda_{gm} = 0.15$ providing a stable starting point. In practice, we recommend first tuning $\rho$ to establish appropriate magnitude balancing, followed by fine-tuning $\lambda_{gm}$ to refine directional alignment \cite{guo2024classifier}.}

\rrevise{Fairness modulation has three key hyperparameters: fairness threshold ($\tau$), fairness penalty ($\lambda_f$), and fairness modulation strength ($\delta$). The fairness threshold ($\tau$) is selected within $[0.0–0.08]$, while both the fairness penalty and fairness strength are chosen within $[0.2–0.8]$. These moderate ranges balance fairness enforcement and predictive performance. We use a protected grid search strategy, varying one hyperparameter at a time while fixing the others at their mid-range values to ensure stable and controlled tuning.}

\color{black}
\subsubsection{Model Training with Multiclass and Missing Modalities}
\label{multiclass_multimodal}
\rrevise{ The complete methodological detail of MultiFair is provided in Section~\ref{multifair_method}. Adapting MultiFair to multiclass tasks with missing modalities requires a few adjustments.
In particular, the surrogate AUC formulation and data loading strategy are carefully adapted to support multiclass optimization while handling demographic fairness constraints under partial modality availability.}

\textbf{Dataloader with Missing Modalities:}
\rrevise{Let the input of sample $j$ consists of $M$ modalities, $\{ x_j^{(1)}, x_j^{(2)}, \dots, x_j^{(M)} \}$. For each modality $m \in \{1,\dots,M\}$, we define a modality availability indicator as:
\[
z_j^{(m)} =
\begin{cases}
1, & \text{if modality } m \text{ is available}, \\
0, & \text{otherwise}.
\end{cases}
\]}

\rrevise{When $z_j^{(m)} = 0$, a zero-valued placeholder tensor matching the spatial dimensions of $x_j^{(m)}$ is inserted to preserve architectural consistency during batched training. To mitigate imbalance across modality combinations, we adopt a modality-aware mini-batch construction strategy that enforces controlled proportions (e.g., $40\%$--$60\%$ via up-sampling) of modality-present samples, thereby ensuring adequate optimization of modality-specific encoders despite limited availability.}

\textbf{Model-Level Masked Fusion:} \rrevise{Let $\{ f_j^{(1)}, f_j^{(2)}, \dots, f_j^{(M)} \}$ denote the learned embeddings from the $M$ modality-specific encoders for sample $j$. The Transformer-based fusion module receives the token sequence $[f_j^{(1)},\, f_j^{(2)},\, \dots,\, f_j^{(M)}]$.
For any modality $m$, when $z_j^{(m)} = 0$, the corresponding embedding is zeroed and a key-padding mask disables attention to its token, such that $\textit{Attention}(f_j^{(m)}) = 0 \quad \textit{if } z_j^{(m)} = 0.$ Thus, the model dynamically operates on the subset of available modalities while maintaining shared parameters across all samples. To avoid biased fairness estimation, surrogate AUC is computed only over samples where all the modalit\revision{ies} are available. Samples with missing modalities are excluded from pairwise comparisons, fairness modulation is skipped if that modality does not appear in a mini-batch, and EMA smoothing is applied to maintain stability. }

\textbf{Multiclass Extension of Surrogate AUC:} \rrevise{For multiclass classification with $K$ classes, we extend the binary surrogate AUC (Algorithm~\ref{alg:surrogate_auc}) using a one-vs-rest strategy. For each class $k \in \{1,\dots,K\}$, the label is converted to a binary label as: \[ y_j^{(k)} = 1(y_j = k).\] 
Here, $y_j^{(k)} =1, \text{if sample } j \text{ belongs to class } k, \text{otherwise 0}.$}
\rrevise{Finally, the class-specific prediction score $s_{j,i,k}$ is used to compute $\text{AUC}^{\text{batch},(t)}_{g,i,k}$ using Algorithm \ref{alg:surrogate_auc}. The multiclass surrogate AUC for subgroup $g$ and modality $i$ is then obtained by macro-averaging:
\[
\text{AUC}^{\text{batch},(t)}_{g,i}
=
\frac{1}{K}
\sum_{k=1}^{K}
\text{AUC}^{\text{batch},(t)}_{g,i,k}.
\]}
\rrevise{This extension maintains differentiability and enables consistent fairness-aware optimization in the multiclass setting. All other components of the MultiFair framework remain unchanged when applied to multiclass classification.
}

\subsection{CheXpert Training and Testing Dataset Insight}
\label{dataset_behavior}
\myrevise{As described in Section \ref{Chexpert_dataset}, the CheXpert dataset contains multiple pathology labels for chest X-ray images with frontal and lateral views. We treat the frontal and lateral views as two modalities, making this a multimodal setting. However, many samples lack the lateral view. During training, we use up-sampling to handle the missing-modality imbalance, so that each batch contains about 40\% paired samples and 60\% single-modality samples, as described in Appendix \ref{multiclass_multimodal}. We use a batch size of 128, and the statistics reported in Table \ref{chexpert_pathology_stats} are consistent with this training setup.}

\myrevise{Table \ref{chexpert_pathology_stats} provides per-pathology subgroup and overall sample statistics for the CheXpert experiments to clarify the stability and coverage of the fairness analysis. For each pathology, the table reports the numbers of male and female training,}  \myrevise{paired-training (frontal and lateral chest x-ray images), and test samples, together with the corresponding subgroup-level fairness triggering rates. It also reports the overall training size, paired-training size, triggering rate, test size, and fallback batch fraction. These statistics clarify the support underlying each pathology-level fairness estimate and indicate how often the fallback mechanism is used, especially for rare pathologies. For example, Pleural Other, Fracture, and Lung Lesion have relatively smaller test support and higher fallback batch fractions, so their fairness estimates should be interpreted more cautiously. Overall, the table helps to demonstrate which pathologies have stronger support and which fairness estimates should be read more carefully because of limited data or higher fallback frequency.}

\begin{table*}[t]
\centering
\caption{\parbox{0.8\textwidth}{\centering
Per-pathology and subgroup statistics for CheXpert dataset. Triggered (\%) denotes the fraction of batches with active fairness modulation, and Fallback Batch Fraction (\%) denotes the fraction of batches where the surrogate AUC was set to 0.5.}}
\label{chexpert_pathology_stats}
\resizebox{0.8\textwidth}{!}{%
\begin{tabular}{lccccccccccccc}
\toprule
& \multicolumn{4}{c}{\textbf{Male}} & \multicolumn{4}{c}{\textbf{Female}} & \multicolumn{5}{c}{\textbf{Overall}} \\
\cmidrule(lr){2-5} \cmidrule(lr){6-9} \cmidrule(lr){10-14}
\makecell{\textbf{Pathology}} 
& \makecell{\textbf{Train}}
& \makecell{\textbf{Train}\\\textbf{Paired}}
& \makecell{\textbf{Test}}
& \makecell{\textbf{Triggered}\\\textbf{(\%)}}
& \makecell{\textbf{Train}}
& \makecell{\textbf{Train}\\\textbf{Paired}}
& \makecell{\textbf{Test}}
& \makecell{\textbf{Triggered}\\\textbf{(\%)}}
& \makecell{\textbf{Train}}
& \makecell{\textbf{Total}\\\textbf{Paired}}
& \makecell{\textbf{Triggered}\\\textbf{(\%)}}
& \makecell{\textbf{Test}}
& \makecell{\textbf{Fallback}\\\textbf{Batch (\%)}} \\
\midrule
No Finding & 11185 & 7229 & 968 & 64.63 & 7674 & 4254 & 717 & 55.43 & 18859 & 11483 & 60.89 & 1685 & 0.23 \\
Enlarg. Cardiomed. & 10953 & 5009 & 1239 & 45.73 & 6782 & 2612 & 757 & 38.51 & 17735 & 7621 & 42.97 & 1996 & 0.46 \\
Cardiomegaly & 15629 & 6068 & 1869 & 38.83 & 10693 & 4013 & 1123 & 37.53 & 26322 & 10081 & 38.30 & 2992 & 0.00 \\
Lung Opacity & 48049 & 15943 & 5699 & 33.18 & 33334 & 10005 & 3965 & 30.01 & 81383 & 25948 & 31.88 & 9664 & 0.00 \\
Lung Lesion & 5150 & 3253 & 475 & 63.17 & 3670 & 1932 & 317 & 52.64 & 8820 & 5185 & 58.79 & 792 & 7.89 \\
Edema & 25355 & 4134 & 3529 & 16.30 & 19475 & 3418 & 2537 & 17.55 & 44830 & 7552 & 16.85 & 6066 & 0.00 \\
Consolidation & 18246 & 6553 & 2112 & 35.91 & 12858 & 4080 & 1436 & 31.73 & 31104 & 10633 & 34.19 & 3548 & 0.00 \\
Pneumonia & 11182 & 5162 & 1167 & 46.16 & 7913 & 3378 & 879 & 42.69 & 19095 & 8540 & 44.72 & 2046 & 0.23 \\
Atelectasis & 29321 & 9826 & 3421 & 33.51 & 19578 & 5798 & 2373 & 29.61 & 48899 & 15624 & 31.95 & 5794 & 0.00 \\
Pneumothorax & 9971 & 3089 & 1249 & 30.98 & 6121 & 1453 & 835 & 23.74 & 16092 & 4542 & 28.23 & 2084 & 0.83 \\
Pleural Effusion & 42499 & 15057 & 5071 & 35.43 & 29364 & 8561 & 3403 & 29.15 & 71863 & 23618 & 32.87 & 8474 & 0.00 \\
Pleural Other & 3399 & 2522 & 276 & 74.20 & 2137 & 1463 & 153 & 68.46 & 5536 & 3985 & 71.98 & 429 & 26.10 \\
Fracture & 4936 & 2464 & 560 & 49.92 & 2657 & 1211 & 287 & 45.58 & 7593 & 3675 & 48.40 & 847 & 15.17 \\
\bottomrule
\end{tabular}%
}
\end{table*}

\color{black}
\subsection{Methodological Justification}

\subsubsection{Analysis for Fairness Activation Rule $\Delta AUC \geq \tau$} 
\begin{figure}
  \centering
    \includegraphics[width=0.49\textwidth]{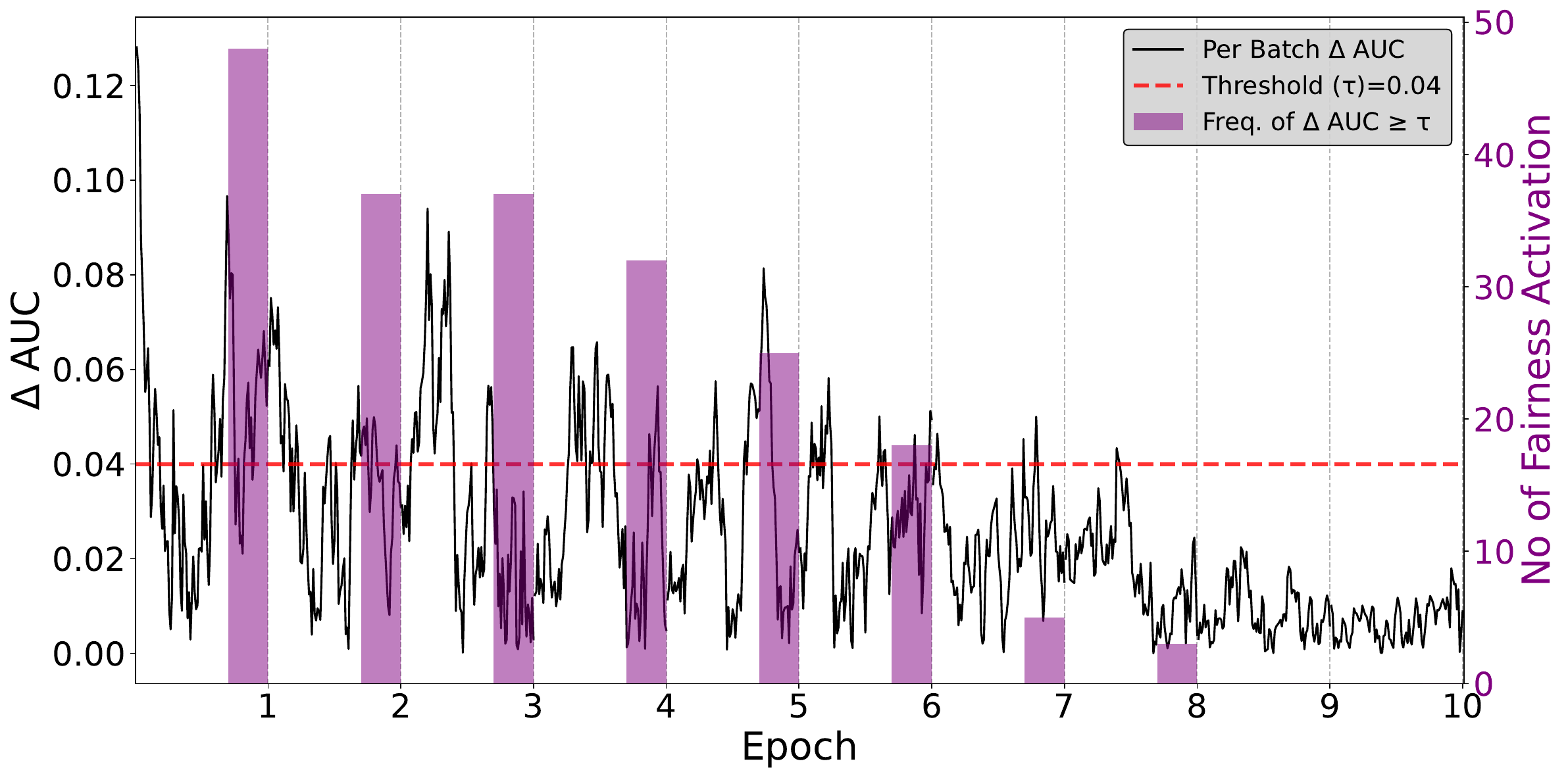}
  \caption{\textbf{Subgroup AUC difference across the batches of each epoch.} Evolution of the subgroup AUC difference ($\Delta AUC$) over training with threshold $\tau = 0.04$; fairness modulation is activated when $\Delta AUC \geq \tau$, and the bars indicate the number of batch-wise fairness activations per epoch for batch size of 64.}
  \label{activation_rule}
  \vspace{-4mm}
\end{figure}

Fig. \ref{activation_rule} plots the per-batch subgroup AUC gap ($\Delta\mathrm{AUC}$)  over training for the MultiFair \revision{framework} with threshold $\tau = 0.04$ on the \rrevise{FairVision \cite{luo2024fairvision} dataset, along with the frequency of batch-level fairness activations per epoch. Early in training, $\Delta\mathrm{AUC}$ shows noticeable short-term fluctuations and frequently exceeds the threshold, leading to repeated but localized fairness activations. As training progresses, both the magnitude and variability of $\Delta\mathrm{AUC}$ steadily decrease, and the number of fairness activations per epoch drops monotonically, indicating that any initial on-off behavior is temporary and does not} \rrevise{persist. Once subgroup performance gaps narrow, $\Delta\mathrm{AUC}$ remains consistently below $\tau$, causing fairness modulation to automatically disengage and demonstrating stable convergence without overshoot or the need for explicit hysteresis.}

\subsubsection{Effectiveness of Dual-Modulation}
\label{dual_modulation}
\begin{figure}[!t]
  \centering
  \subfloat[]{
    \includegraphics[width=0.48\linewidth]{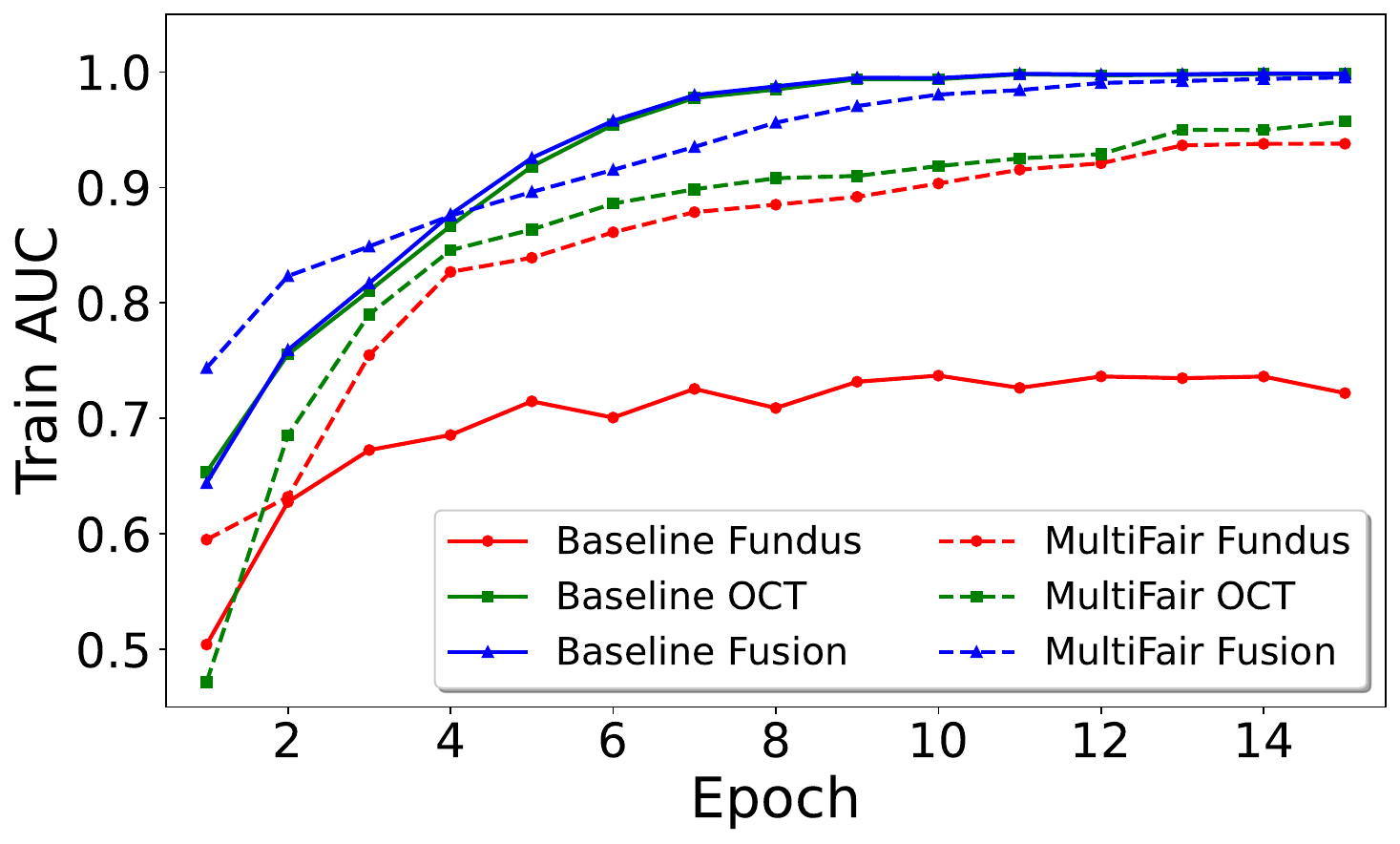}\label{modality}}
  \hfill
  \subfloat[]{
    \includegraphics[width=0.48\linewidth]{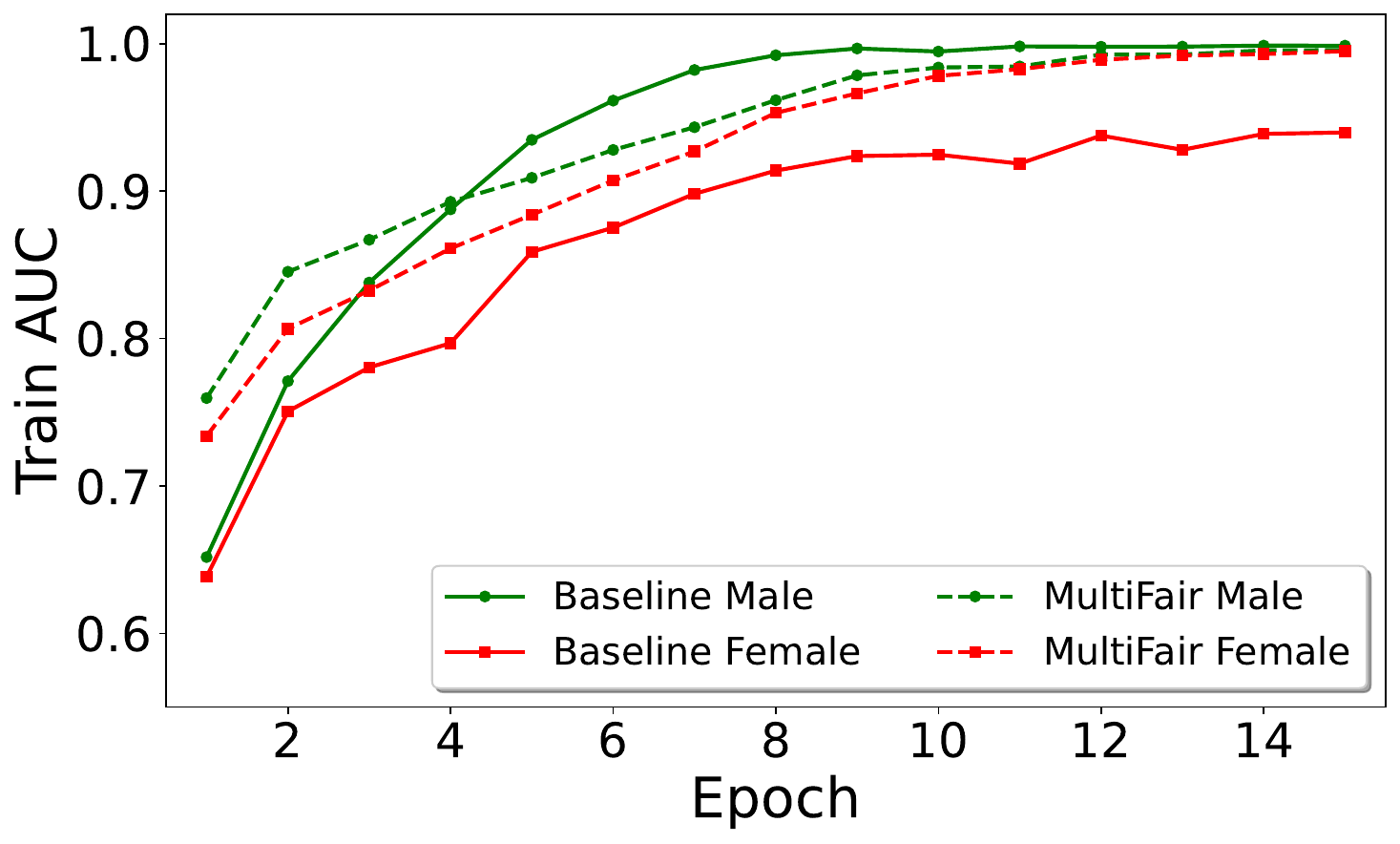}\label{subgroup}}
    \vspace{-1mm}
  \caption{\textcolor{black}{\textbf{Modality and subgroup modulation during training.} Training time modulation effect for 15 epochs on the FairVision dataset with batch size of 32.The figure compares MultiFair and the baseline model (without modulation), illustrating AUC trends during training for (a) modality-level modulation and (b) subgroup-level modulation.  }}
  \label{modulation_training}
  \vspace{-5mm}
\end{figure}

\rrevise{We theoretically proved how MultiFair \revision{balances} the modality and subgroups in the section \ref{optimization}.  In figure \ref{modulation_training}, we empirically shows how MultiFair promotes balanced learning at both} \rrevise{modality (Fig. \ref{modality}) and subgroup (Fig. \ref{subgroup}) levels through adaptive modulation during training. At the} \rrevise{modality level, it prevents dominance of stronger modalities (e.g., OCT) by scaling gradients to ensure weaker modalities (e.g., fundus) continue to improve, leading to more aligned AUC trends across modalities. At the subgroup level, MultiFair reduces the performance gap between demographic groups by limiting updates that overly favor advantaged groups. As seen in the training curves (Fig. \ref{subgroup}), this leads to more balanced and stable AUC improvements across groups, helping the model learn more equitably without reducing overall performance.} 

\rrevise{However, we see in Fig. \ref{modality} that the modalities have not converged exactly to zero disparities. Because stochastic optimization, subgroup imbalance, non-convexity, and gradient interactions may prevent strict equality. We also empirically observe that after approximately five to six epochs, validation performance begins to overfit even though modality imbalance} \rrevise{and subgroup gaps continue decreasing, as shown in Figure \ref{modulation_training}. So, we use early stopping to maintain balanced predictive and fairness performance. Thus, we ensure trade-off between fairness optimization and generalization.}

\begin{figure}[!t]
  \centering
  \subfloat[]{
    \includegraphics[width=0.45\linewidth]{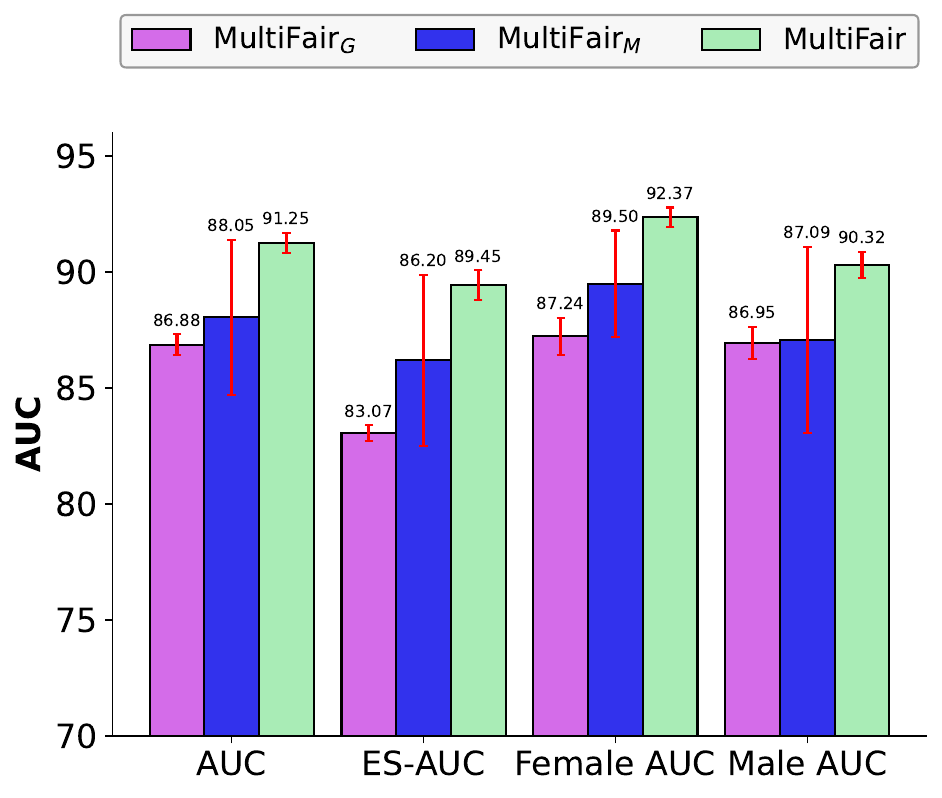}\label{gender_fairclip}}
  \hfill
  \subfloat[]{
    \includegraphics[width=0.52\linewidth]{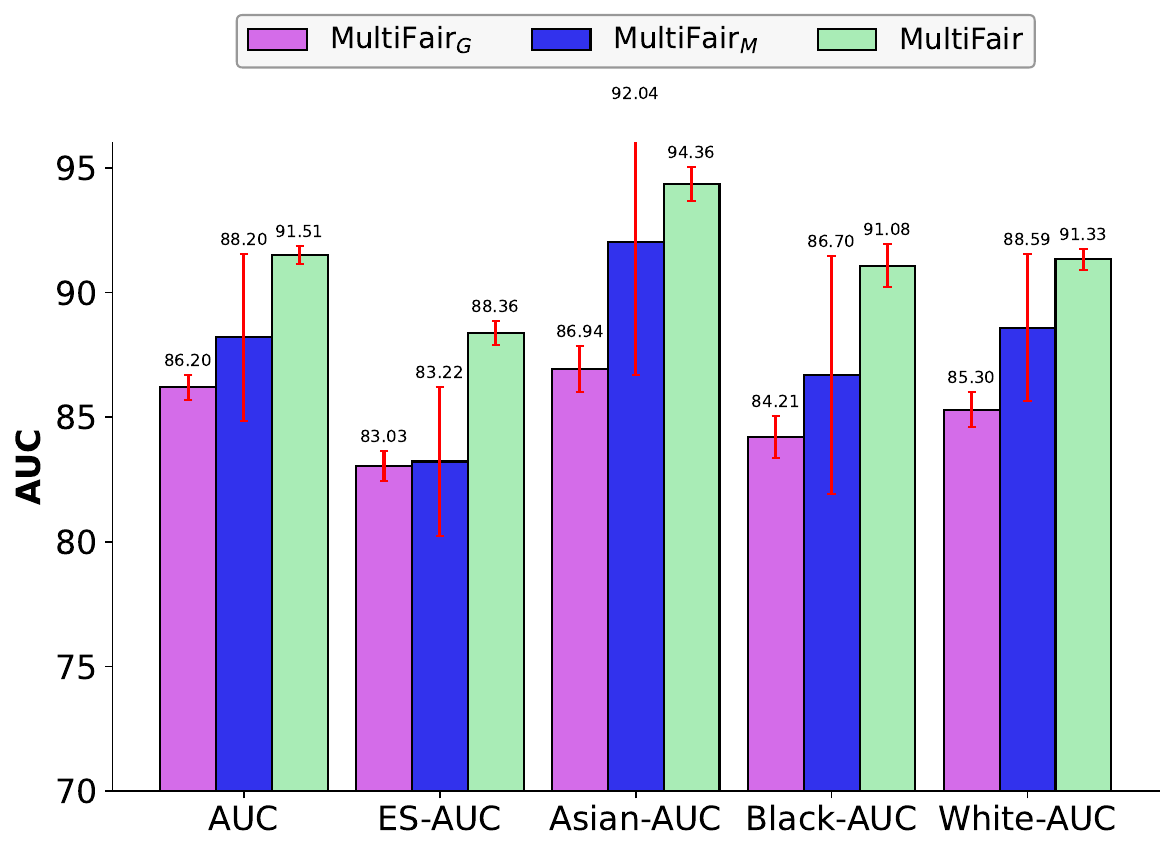}\label{race_fairclip}}
    \vspace{-1mm}
    \caption{\rrevise{\textbf{Additional Ablation study results on FairCLIP dataset}. We represent the performance of fairness only with \text{\modelname}$_G$, modality only with \text{\modelname}$_M$, and modulation only with \text{\modelname} for the MultiFair framework. The figure (a) represents the performance for the gender groups, and (b) shows the corresponding performance for various racial subgroups.}}
  \label{fig_ablation_add}
  \vspace{-3mm}
\end{figure}

\begin{figure}[!t]
  \centering
    \includegraphics[width=0.7\linewidth]{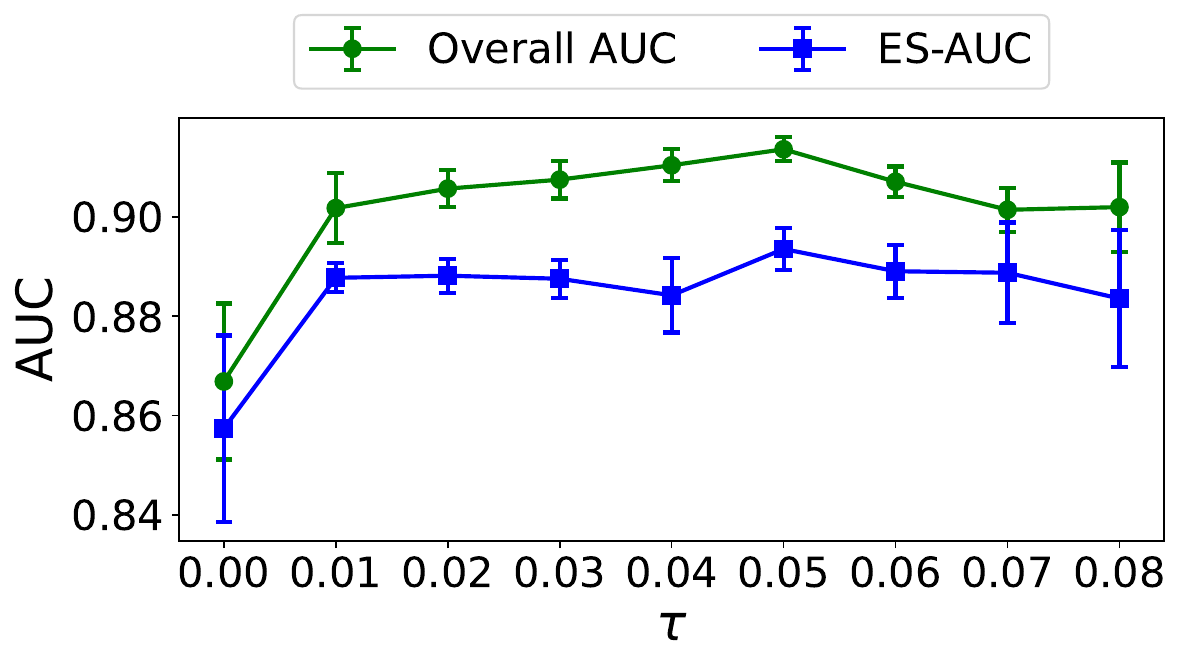}\label{tau_fairclip}
  \caption{\rrevise{\textbf{The impact of different fairness threshold on FairCLIP Dataset}. The horizontal lines indicate the mean AUC values, while the vertical bars represent the corresponding standard deviations.} } 
  \label{fairness_trigger}
  \vspace{-5mm}
\end{figure}

\color{black}
\subsection{Additional Ablation Study}
\label{additional_ablation}
\rrevise{To further investigate the interaction between modality balancing and fairness-aware modulation, we analyze the ablation results shown in Fig. \ref{fig_ablation_add} over 6 random independent experiments ($Mean\pm Standard \ Deviation$) on the FairCLIP dataset. The comparison between $\text{MultiFair}_{G}$ (fairness-only), $\text{MultiFair}_{M}$ (modality-only), and the full MultiFair model demonstrates that combining the two mechanisms consistently yields the best overall performance and subgroup metrics. Specifically, while $\text{MultiFair}_{G}$ improves subgroup parity by directly reducing disparities, it does not fully leverage } \rrevise{complementary modality information. Conversely, $\text{MultiFair}_{M}$ improves representation quality by balancing modality gradients but does not explicitly address subgroup disparities. The full MultiFair model integrates both mechanisms, allowing the optimization process to simultaneously correct subgroup imbalance and maintain balanced modality contributions. This joint modulation leads to improved overall AUC and more consistent subgroup performance across both gender and racial groups, suggesting that the two components address complementary aspects of the learning problem.}

\rrevise{We also evaluate the stability of the framework under different fairness triggers by varying the fairness thresholds $\tau$, as shown in Fig.~\ref{fairness_trigger} (3 random independent experiments). The results indicate that the model remains stable across a wide range of threshold values. In particular, both the overall AUC and ES-AUC remain relatively consistent for $\tau$ values between 0.01 and 0.08, with only minor variations across runs. This behavior indicates that the fairness modulation is not overly sensitive to the choice of threshold and that the framework maintains a favorable balance between predictive performance and fairness objectives.}

\section*{Acknowledgment}
\rrevise{The authors would like to thank Dr.Nussdorf Jonathan for providing valuable clinical insights during the preparation of this paper.}

\bibliographystyle{IEEEtran}
\color{black}
\bibliography{references_r2_clean}

\end{document}